\documentclass{article}
\usepackage{graphicx} 
\usepackage{mathtools}
\usepackage{amsmath, amssymb}
\usepackage{algorithm}
\usepackage{algpseudocode}
\usepackage[utf8]{inputenc}
\usepackage{hyperref}
\usepackage{subfig}
\usepackage[toc,page]{appendix}
\usepackage{abstract}
\usepackage[english]{babel}
\usepackage[a4paper, total={6in, 8in}]{geometry}
\newtheorem{theorem}{Theorem}
\newtheorem{corollary}{Corollary}[theorem]
\newtheorem{lemma}{Lemma}
\newtheorem{assumption}{Assumption}

\newcommand\numberthis{\addtocounter{equation}{1}\tag{\theequation}}

\DeclareMathOperator{\E}{\mathbb{E}}

\title{\textbf{Ravnest: Decentralized Asynchronous Training on Heterogeneous Devices}}

\author{Anirudh Rajiv Menon$^\dagger$ \and Unnikrishnan Menon$^\dagger$ \and Kailash Ahirwar$^\dagger$}
\date{%
    $^\dagger$\textit{Raven Protocol}\\%
}

\begin{document}

\maketitle

\begin{abstract}

Modern deep learning models, growing larger and more complex, have demonstrated exceptional generalization and accuracy due to training on huge datasets. This trend is expected to continue. However, the increasing size of these models poses challenges in training, as traditional centralized methods are limited by memory constraints at such scales. This paper proposes an asynchronous decentralized training paradigm for large modern deep learning models that harnesses the compute power of regular heterogeneous PCs with limited resources connected across the internet to achieve favourable performance metrics. Ravnest facilitates decentralized training by efficiently organizing compute nodes into clusters with similar data transfer rates and compute capabilities, without necessitating that each node hosts the entire model. These clusters engage in $\textit{Zero-Bubble Asynchronous Model Parallel}$ training, and a $\textit{Parallel Multi-Ring All-Reduce}$ method is employed to effectively execute global parameter averaging across all clusters. We have framed our asynchronous SGD loss function as a block structured optimization problem with delayed updates and derived an optimal convergence rate of $O\left(\frac{1}{\sqrt{K}}\right)$. We further discuss linear speedup with respect to the number of participating clusters and the bound on the staleness parameter. 

\end{abstract}

\section{Introduction}

Large Language Models, Diffusion based models and various large scale multi modal architectures  \cite{wei2022emergent, chang2023survey, alayrac2022flamingo, wu2023next, croitoru2023diffusion, yang2023diffusion} becoming the zeitgeist in recent years have created growing requirements for hardware than can handle their training. With the number of model parameters going into the billions and data-sets ranging in terabytes of space \cite{villalobos2022machine, shoeybi2019megatron}, it has become infeasible for an individual system to conduct the entire training process \cite{mattson2020mlperf}. 

Towards distributing the training process across multiple systems, Data Parallelism has been a very effective mechanism in which each system is responsible for a portion of the training data batch \cite{You2020Large, goyal2017accurate}. In Synchronous Data Parallelism, the workers have to wait for gradients to be computed across all peers so that parameter updates can take place and the next data batch can be processed. However, the presence of workers that are much slower than others will significantly impede efficiency. To overcome this, asynchronous mechanisms were introduced where peers don't have to wait for each other to make parameter updates \cite{xu2021dp}. Though asynchronous mechanisms may give rise to updates with stale gradients, they have been observed to converge efficiently under certain constraints \cite{dutta2021slow}. In distributed training setups, communication-efficient All-Reduce mechanisms, such as the widely-used Nvidia Collective Communications Library (NCCL) \cite{nvidiaNVIDIACollective, tanaka2018large}, are preferred for data transmission between peers. This approach is favored over a common parameter server, which can become a bottleneck when the number of communicating peers is significantly large.

Despite its advantages, data parallelism still requires each system to host the entire model which may not always be practicable. Model parallelism helps alleviate this by dividing the model across multiple workers which communicate with each other to conduct forward and backward passes \cite{brown2020language, zhuang2023optimizing, huang2019gpipe}. Traditional model parallelism requires each worker to wait for data from its preceding or succeeding peers in the computation graph before proceeding with a forward or backward pass respectively. This not only leads to bubbles during which multiple workers remain idle but is also vulnerable to stragglers that can hold up the training process. 

This paper details an asynchronous decentralized parallel training approach that utilizes both data and model parallelism, allowing weaker systems to participate in distributed training while minimizing idle time bubble. The proposed method eliminates the need for workers to host entire models and mitigates the reduction in efficiency due to stragglers by grouping workers into clusters within which \textit{Zero-Bubble Asynchronous Model Parallelism} is executed. Periodic parameter synchronisation across clusters is conducted via a communication efficient \textit{Parallel Multi-Ring All-Reduce} mechanism. Theoretical analysis and derivations for convergence rate in terms of the number of updates (\(K\)) show consistencies with local SGD. We also discuss linear speedup with respect to the number of participating clusters and constraints on the delay parameter under which this can be achieved. This mechanism can efficiently train large state of the art models while significantly reducing hardware requirements and training costs, opening up new possibilities for groundbreaking research and real-world applications.

The main contributions of this paper are summarised as follows:
\begin{itemize}
    \item Ravnest, a novel asynchronous parallel training approach that amalgamates the best features of data and model parallelism to facilitate the distributed training of complex deep learning models over large datasets on clusters of heterogeneous consumer grade PCs over the internet, is introduced. 
    \item A Parallel Multi-Ring approach which improves on single Ring All-Reduce has been utilized for parameter averaging. 
    \item Under realistic assumptions, the proposed algorithm is outlined and  and a theoretical analysis which includes derivations for convergence rate of \(O\left(\frac{1}{\sqrt{K}}\right)\), which is consistent with local SGD, is provided. Constraints under which the algorithm can observe linear speedup are detailed.
    \item Our experiments show significant memory reductions and competitive convergence rates with conventional synchronous training, complementing the theoretical analysis. 
\end{itemize}

\section{Related Works}

\subsection{Data and Model Parallelism} 
Due to exponential growth in dataset and model sizes, the need for parallelization and memory usage reduction led to the development of data parallelism. This technique involves using multiple systems, each hosting copies of the entire model, to process portions of a larger data batch in parallel followed by gradient aggregation and parameter updates \cite{shallue2019measuring, valiant1990bridging, li2020pytorch}. This approach presents some advantages such as smaller batch sizes per system, full utilization of compute resources, faster training times \cite{zhao2023pytorch}, scalable infrastructures and support for synchronous and asynchronous updates based on the problem requirements. However, it is evident that on weaker systems with limited memory resources, hosting the entire model may not necessarily be feasible. 
\begin{figure}[htp]
  \centering
  \subfloat[Model Parallelism]{\includegraphics[width=0.35\textwidth]{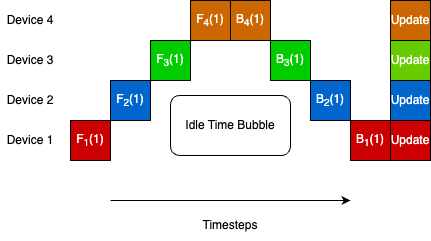}\label{fig:f1a}}
  \hfill
  \subfloat[Pipeline Parallelism]{\includegraphics[width=0.55\textwidth]{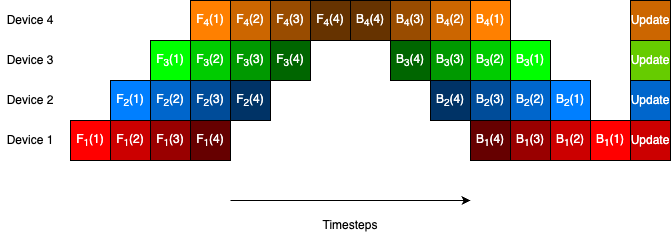}\label{fig:f1b}}
  \caption{Illustration of how pipeline parallelism reduces idle time bubble \cite{huang2019gpipe}}.
\end{figure}

Recent research has introduced model parallelism, a decentralized training paradigm enabling large model training across multiple devices, with each device holding only a portion of the main model it can store \cite{yang2021pipemare, narayanan2021memory, guan2019xpipe}. A pipeline of GPU devices is created with each device hosting a submodel processing data batches and passing results to the next device for further operations. Network traffic and compute demand are key factors for scalability. However, this sequential nature leads to under-utilization of compute resources since there occurs an idle bubble of timesteps as can be observed in Figure \ref{fig:f1a}. To address the issue, pipeline parallelism divides mini-batches into smaller micro-batches, processing them simultaneously across multiple devices (as shown in Figure \ref{fig:f1b}), thereby reducing the idle time bubble and resulting in almost linear speedup \cite{huang2019gpipe}. Pipeline parallelism inherently requires the participating devices to regularly communicate micro-batch results with each other. Recent works have improved communication efficiency by overlapping an optimized cross-mesh resharding technique with the computations on the workers, thereby reducing latency \cite{zhuang2023optimizing}. Asynchronously performing the forward and backward jobs at each pipeline stage while maintaining a balance between communication and computation can significantly reduce communication overhead and idle bubble time \cite{harlap2018pipedream}. This has been achieved by scheduling the forward and backward passes in a round robin fashion. Significant reduction in idle time bubble has been successfully achieved in CNN architectures by considering varying degrees of staleness for each layer \cite{xu2020acceleration}.

\subsection{Asynchronous Stochastic Gradient Descent}

Initially, data parallelism relied on Synchronous Stochastic Gradient Descent (SSGD), which periodically synchronized all workers' gradients before each parameter update. However, this caused delays as workers waited for each other, slowing training and increasing idle time \cite{cipar2013solving}. Asynchronous Stochastic Gradient Descent (ASGD) lets workers train independently by accepting updates with outdated gradients while minimizing wait times caused by straggling peers. Enhanced versions include adjusted averaging mechanisms and approximation functions to handle staleness, enabling ASGD to rival synchronous methods in non-convex optimization problems \cite{zheng2017asynchronous, harlap2018pipedream, xu2020acceleration}. Distributed training traditionally relied on workers communicating with a dedicated parameter server for gradient aggregation and weight updates \cite{li2014scaling, recht2011hogwild, zhang2015staleness, zhao2020distributed}. However, bottleneck and scalability issues arise since the parameter server has to transmit large model-sized payloads to all workers regularly \cite{elgabli2020communication}. Workers update their local models by communicating with peers, often using all-reduce mechanisms for efficient averaging, which are preferred over gossip-based methods due to their efficiency with more peers \cite{dai2023efficient, scaman2019optimal, uribe2020dual}. Reduction mechanisms are also employed to limit worker communication, spanning multiple rounds to minimize network traffic \cite{cui2021asynchronous, ryabinin2021moshpit, ueno2019exhaustive}.

\subsection{Reduction Techniques}
\label{reductiontechniques}

\begin{figure}[h!]
    \centering    \includegraphics[width=\textwidth]{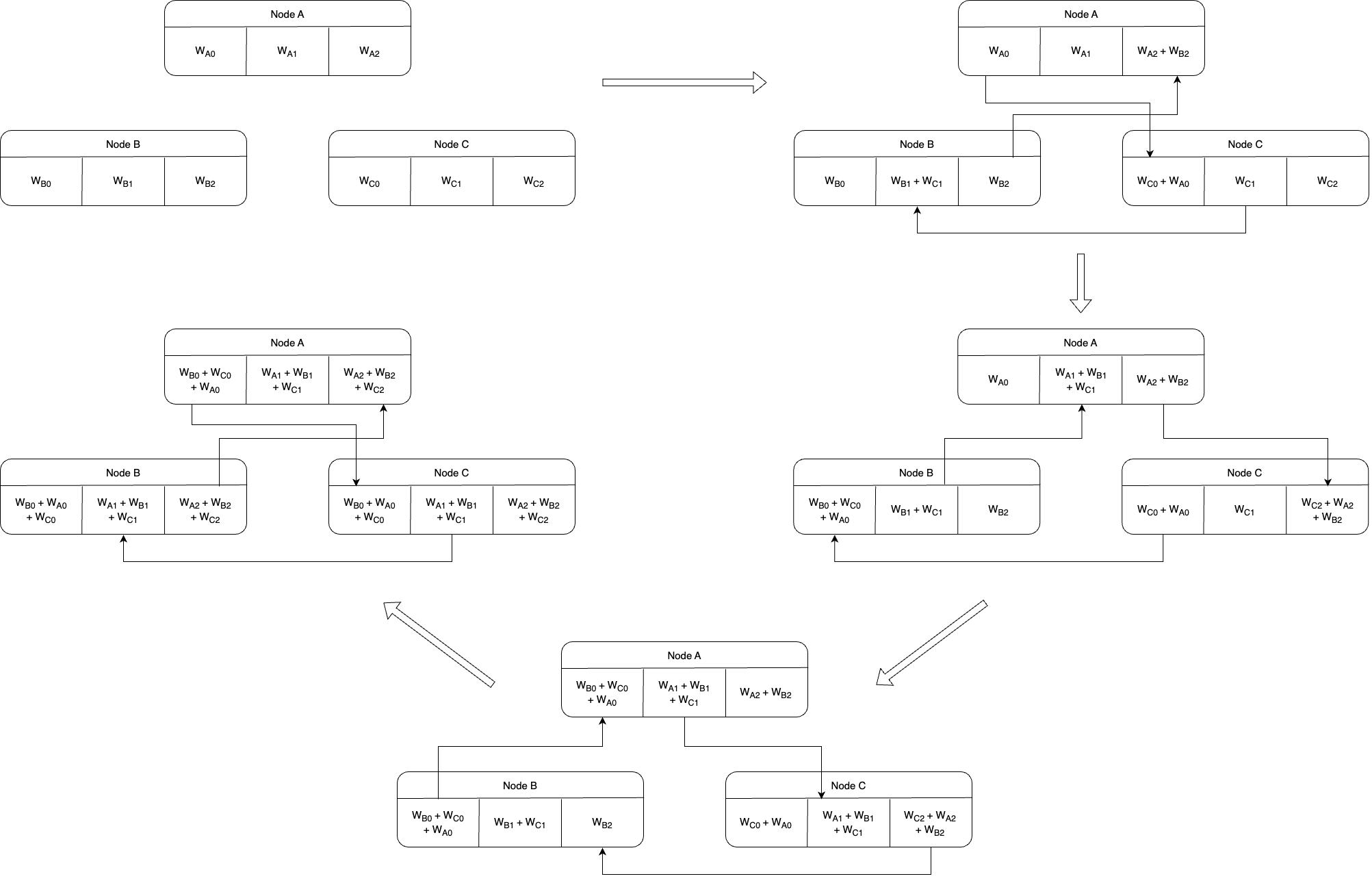}
    \caption{One cycle of Ring All-Reduce across 3 nodes}
    \label{fig:ring}
\end{figure}

In parallel computing, all-reduce algorithms are responsible for combining and redistributing the results originating from multiple processes. One of the most popular approach to achieve this has been the Butterfly All-Reduce algorithm \cite{rabenseifner2004optimization, patarasuk2009bandwidth}. This technique primarily uses a recursively halving all-scatter operation followed by a recursively doubling all-gather operation \cite{wei2021deploying}. This technique is well suited in scenarios where network contention issues don't generally arise. It minimises the number of communication rounds while ensuring each node only shares the minimum amount of data required for reduction. However, in widely deployed clusters, the butterfly all-reduce algorithm bottlenecks due to high network contention. These issues led to more research on alternate reduction techniques like Ring All-Reduce algorithm which requires comparatively lower working memory and has significantly outperformed other algorithms for sufficiently large sizes of data \cite{patarasuk2009bandwidth}. One of the most popular implementations of the Ring All-Reduce algorithm is Uber's Horovord framework, which adds a layer of All-Reduce based MPI training to Tensorflow. \cite{sergeev2018horovod}. Figure \ref{fig:ring} illustrates how $3$ nodes can perform $4$ rounds ($1$ cycle) of Ring All-Reduce to aggregate their weights.

Consider a system of $N$ devices that are connected under a ring topology. Let $i \in \left[1,2,...,N\right]$ represent the ID assigned to each worker in this system increasing in clockwise direction. The main dataset (of size $D$) is first split into $N$ chunks, each of which are distributed evenly across all devices. This means each device holds a mini dataset of size $D/N$. One cycle of Ring All-Reduce comprises $2*(N-1)$ rounds of sharing among these devices. In each of these rounds, device $i$ sends its parameters to device $(i+1)\mod N$. As and when a device receives the parameters from the preceding device, they apply the reduce operator on it and a new set of averaged parameters are calculated and shared with the subsequent device in the ring. These new averaged parameters keep getting propagated in clockwise direction $2*(N-1)$ times, after which all $N$ devices will be having a common set of averaged parameters. Each device reduces $O(D/N)$ data at a time but has to communicate with only its immediate left and right neighbours. While Ring All-Reduce beats most approaches in terms of communication cost, latency issues arise when the number of participating nodes is very large. 

\subsection{Training over the Internet}

Advancements in Natural Language Processing, Computer Vision and other domains of artificial intelligence have shown us that training models with an increasingly larger number of parameters with all the more complex architectures on gigantic datasets yields systems that can make extremely accurate predictions. The caveat is that efficiently training these massive models requires the spending of millions of dollars in setting up high-end GPU server pods (each containing thousands of GPUs) that are interconnected with low latency links in specialized environments. 

This hinders breakthroughs in the field as researchers without access to such high-computing systems are left with no other options. This restricts competitive research and making breakthroughs in the field of AI exclusively to organizations with massive funds and hardware resources \cite{diskin2021distributed}. In the future, as models get more and more complex, this approach might not be scalable and sustainable. Combating these constraints led to more extensive research on how regular desktop systems and laptops with varying specifications can be pooled together to train large models over the internet \cite{ryabinin2020towards}. Compared to the high-performance computing accelerators, these individual PCs are slow and unreliable. However, paradigms that make use of these PCs can lay the foundation for the future of AI training as it democratizes research among those with limited resources at the cost of higher training times. An interesting application of this approach of pooling compute nodes across the internet can be seen in the project Leela Chess Zero \cite{leelazero}. It relies on volunteer compute power to generate self play data for chess games and uses it to train reinforcement learning agents. 
 
\section{Proposed Method}

We present a communication efficient method for asynchronous decentralised model parallel SGD. The proposed algorithm aims to facilitate distributed learning across a network of nodes without depending on a parameter server setup and each node having to operate on only a portion of the entire model.

We aim to solve the following block structured optimization problem of minimizing the loss function:

 \begin{align*} \min_x f(x) = \dfrac{1}{C}\sum_{i=1}^{C} f_i(x) \numberthis \label{eqn1}\end{align*}

where $C$ is the number of clusters, $f_i$ is the loss function associated with the $i^{th}$ cluster having $n_i$ disjoint blocks such that $x = (x_{i,1}, x_{i,2},....,x_{i,n_i})$ on the $i^{th}$ cluster, where $x_{i,j}$ represents the parameters of submodel $j$. 

The algorithm involves assigning participating nodes into clusters within which they will further be assigned a submodel to participate in \textit{Zero-Bubble Asynchronous Model Parallel} training. The clusters will also periodically communicate with each other to conduct a global averaging of parameters. The following sections detail the various procedures involved.

\subsection{Zero-Bubble Asynchronous Model Parallel Training}

Within each cluster, an Asynchronous Model Parallel procedure is used which eliminates the idle time bubble observed in traditional Model Parallel techniques. 

We can view each cluster as a computation graph where each peer (containing a submodel) forms a node and communicates with a subset of peers in order to perform forward and backward passes. The submodel on each peer asynchronously conducts forward and backward passes as and when it receives data from its preceding or succeeding peers in the computation graph as seen in Algorithm \ref{alg:cap1}.

\begin{algorithm}[htp]
\caption{Zero-Bubble Asynchronous Model Parallelism Within A Cluster}\label{alg:cap1}
\begin{algorithmic}[1]
\Require cluster id: $c$, cluster model parameters: $x_c$, learning rate: $\eta$
\While{training is not terminated}:
    \For{each peer $p$ in cluster running parallelly}
        \If{gradients received in backward-buffer}
            \State Pop gradients from backward-buffer 
            \State Calculate update using gradients from backward-buffer: $x_{c,p} \leftarrow x_{c,p} - \eta \hat{g}_{c,p}$
            \State Propagate required gradients to preceding peers in Cluster Graph  
        \EndIf
        \If{activations received in forward-buffer}
            \State Pop activations from forward-buffer
            \State $O = forward\_pass(x_{c,p}, \text{forward-buffer values})$
            \State Forward activations $O$ to succeeding peers in Cluster Graph
        \EndIf
    \EndFor
\EndWhile
\end{algorithmic}
\Comment{Blocks 3-7 and 8-12 can be run in parallel with the communication process using Process Locks. More details are available in the Implementation Details section} 
\end{algorithm}

The submodels don't wait for gradients of the current data mini-batch from the succeeding peer before proceeding with the forward pass for the next one. When a submodel receives back-propagated gradients from its succeeding nodes, it conducts the update step with respect to its parameters immediately as per the following equation: 

\begin{align*}
x_{c,p}^{k+1} = x_{c,p}^k - \eta \hat{g}_{c,p}^k
\numberthis \label{eqn2}
\end{align*}

where $\hat{g}_{c,p}^k = g_{c,p}^{k-\tau_k} = \nabla_p F(x_c^{k-\tau_k}, b_c^k)$, where $F$ denotes the loss function utilized in the training process, and $x_c^{k-\tau_k}$ is used to denote the state of the model parameters when the forward pass of batch $b_c^{t}$ was being conducted, depicting that the gradient being used for the update is stale as the submodel may have undergone updates during the intermediate time-steps. 

\begin{figure}[htp]
    \centering    \includegraphics[width=0.5\textwidth]{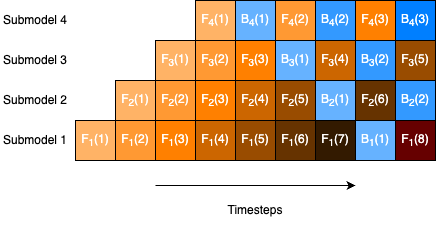}
    \caption{Zero-Bubble Asynchronous Model Parallelism within one cluster}
    \label{fig:zerobubble}
\end{figure}

It can be seen from Figure \ref{fig:zerobubble} that the degree of staleness $\tau_k$ increases as we move towards peers containing the initial parameters of the model. Once backward pass and parameter updates have been conducted the required gradients are then propagated to the preceding node. Due to the asynchronous nature of the forward and backward passes, the idle time bubble is reduced to zero. Note that the degree of staleness decreases in the latter nodes, reducing to zero in the final edge node.

\subsection{Parallel Multi-Ring All-Reduce Mechanism for Parameter Averaging}
\label{parallelmultiring}

Ravnest runs multiple Ring All-Reduces simultaneously to parallelize the averaging of model parameters across all clusters. 

\begin{figure}[htp]
    \centering    \includegraphics[width=0.6\textwidth]{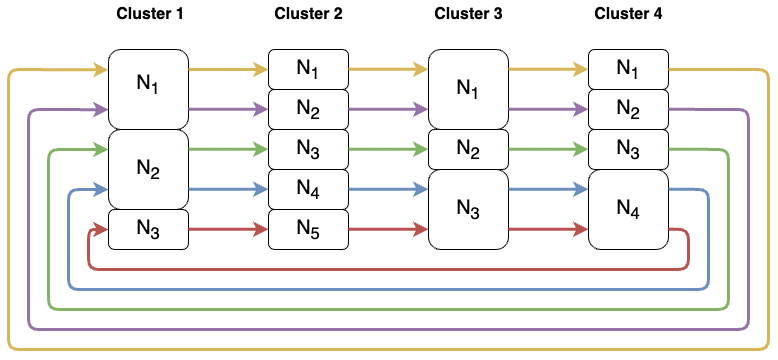}
    \caption{One round of Parallel Multi-Ring All-Reduce, with 5 parallel rings, during global parameter averaging.}
    \label{fig:parallelring}
\end{figure}

As discussed in section \ref{reductiontechniques}, one cycle of ring all-reduce comprises of $2*(C-1)$ rounds for $C$ clusters. Figure \ref{fig:parallelring} illustrates one round of this parallel reduction procedure across 4 heterogeneous clusters, each containing a different number of peers. Note that the area of each node $(N_1, N_2, ...)$ is proportional to the size of the submodel hosted on it. For instance,
$size(Cluster2:N_1)$ is equivalent to $size(Cluster1:N_1) + size(Cluster1:N_2)$. 

Before training commences, our cluster formation approach (discussed in section \ref{clusterformation}) optimally groups together nodes into clusters. Each peer is also provided necessary information for the reduction procedure in the form of mappings that indicate the corresponding workers from other clusters that it needs to form a ring with so that its submodel parameters can participate in averaging. Within a cluster, since peers contain submodels of varying sizes, a node may have to be a part of multiple rings and share/receive the required submodel parameters with one or more peers of the neighbouring clusters. This also implies that the number of rings that will be formed is equivalent to the number of maximum nodes across all clusters. 

All peers of a cluster can now simultaneously send over their respective chunks of model parameters for aggregation to the responsible peers in the successor cluster. This marks one round of our Parallelized Multi Ring All-Reduce. Once received, each peer aggregates their existing parameters with the received parameters and keeps it ready for the next round. $2*(C-1)$ such rounds make up one cycle of parameter aggregation post which, all clusters effectively should be having the averaged model parameters.  

The complexity of each parallel ring is $O\left(2*(C-1) * \dfrac{S}{C}\right)$, $S$ being the size of the parameters being averaged in that ring. The complexity for one entire all-reduce process can thus be represented as $O\left(2*(C-1) * \dfrac{S_{\text{max}}}{C}\right)$ where $S_{\text{max}}$ is the maximum data size being averaged across all the rings. Since this value will be a fraction of the entire model's size and each peer is averaging only a portion of a submodel at a time, this mechanism performs favourably compared to having a single ring for the entire model which would face either high latency issues or large bandwidth usage depending on the number of participating workers.

\subsection{Ravnest Algorithm}

The participating nodes in a training session are first grouped together into clusters based on their individual RAM and bandwidth specifications such that no cluster is significantly more capable than the others as shown in section \ref{clusterformation}, following which each node is assigned a submodel that it is capable of training.

\begin{algorithm}[h!]
\caption{Ravnest Algorithm}\label{alg:cap2}
\begin{algorithmic}[1]
\Require common initial weight for all clusters: $x^0$, learning rate: $\eta$, communication period $\kappa$
\For{$t = 0,1,...$}
    \State Randomly Sample a cluster $i_t$ from $\left[C\right]$
    \State Randomly Sample a peer $p_{i_t,t} \in$ cluster $i_t$
    \If{$t+1 \mod \kappa = 0$}
        \State $x_{i}^{t+1} = \text{Parallel Multi-Ring All-Reduce}(x_{1}^{t}, x_{2}^{t},..., x_{i_t}^{t} - \eta S_{p_{i_t,t}} \hat{g}_{i_t}^t,...,x_{C}^{t}) \quad \forall i \in [C]$
    \Else
        \State $x_{i_t, p_{i_t,t}}^{t+1} = x_{i_t, p_{i_t,t}}^{t} - \eta \hat{g}_{i_t,p_{i_t,t}}^t$
    \EndIf
\EndFor
\end{algorithmic}
\Comment{Here each peer participates in \textit{Zero-Bubble Asynchronous Model Parallelism} within their cluster and conducts updates as shown in Algorithm \ref{alg:cap1}} 
\end{algorithm} 

Each cluster can effectively be viewed as a graph of interconnected nodes wherein each individual node is responsible for a portion of the model, i.e. the parameters on the $i^{th}$ cluster can be represented as $x_i = (x_{i,1}, x_{i,2},...x_{i,P_i})$ where $P_i$ is the number of submodels/nodes on the $i^{th}$ cluster. Note that the minimum number of submodels that a cluster can have is denoted by $N_m$.

In Algorithm \ref{alg:cap2}, all workers across all clusters run parallelly in an asynchronous fashion and \textit{Zero-Bubble Asynchronous Model Parallelism} is used within each cluster as shown in Algorithm \ref{alg:cap1} and updates are made as and when Gradients are received by a peer $p_i$ in cluster $i$: 

\begin{align*}
x_{i, p_{i}}^{t+1} = x_{i, p_{i}}^t - \eta \hat{g}_{i,p_{i}}^t
\numberthis \label{eqn3}
\end{align*}

where $x_{i, p_{i}}^{t+1}$ represents the parameters of peer $p_{i}$ on cluster $i$ and $\hat{g}_{i,p_{i}}^t = \nabla_{p_i} F(x_{i}^{t-\tau_t}, b_i^t)$

Model parameters are also averaged periodically across clusters using the reduction mechanism detailed in section \ref{parallelmultiring}. Since at any given time-step a submodel on a peer updates its weights, we represent these updates in terms of the parameters at the cluster level as follows:

\begin{align*}
x_{i}^{t+1} = x_{i}^{t} - \eta S_{p_{i}} \hat{g}_i^t \numberthis \label{eqn4}
\end{align*}

$x_i$ being the model parameters on cluster $i$, $\hat{g}_i^t = \nabla F(x_{i}^{t-\tau_t}, b_i^t)$ and $S_{p_{i}}$ is a sampling matrix such that for a vector $v$ with $v_i$ as its $i^{th}$ block, $S_iv$ has $v_i$ in its $i^{th}$ block and $0$ in the others.  

Note that $t$ is a global iteration counter and increments whenever an update is performed by any peer on any cluster. This can be interpreted such that at any time-step $t$ a peer $p_{i_t,t}$ on cluster $i_t$ updates its parameters and increments $t$. 

The convergence rate of the proposed mechanism has been explored in section \ref{theoreticalanalysis}.

\subsection{Theoretical Analysis}
\label{theoreticalanalysis}
We explore the convergence of the proposed method in this section. Towards this, we use the following assumptions in our theoretical analysis, proofs and derivations.

\begin{assumption}
\label{Assumptions}

\item[1.] \textbf{Unbiased Gradient:}
We assume that the data has been distributed appropriately allowing for the stochastic gradients to be unbiased:
\begin{align*}
\E_{b_i}\left[\nabla F(x, b_i)\right] = \nabla f_i(x)
\numberthis \label{eqn5}
\end{align*}

where $b_i$ is a data minibatch from the data belonging to $i^{th}$ cluster, and
\begin{align*}
\E_i \left[ f_i(x) \right] = f(x)
\numberthis \label{eqn6}
\end{align*}

\item[2.] \textbf{Lipschitz condition:}
All $f_i$'s are $L-smooth$ and their gradients are Lipschitzian, i.e.
\begin{align*}
\left\Vert \nabla f_i(y) - \nabla f_i(x) \right\Vert \leq  L \left\Vert y-x \right\Vert
\numberthis \label{eqn7}
\end{align*}

and
\begin{align*}
f(y) \leq f(x) + \left\langle \nabla f(x), y-x \right\rangle + \dfrac{L}{2} \left\Vert y-x \right\Vert^2
\numberthis \label{eqn8}
\end{align*}

\item[3.] \textbf{Bounded Variance:}
The variance of the stochastic gradient is bounded for all $i$'s:
\begin{align*}
\E_{b_i}\left[ \left\Vert \nabla F(x, b_i) - \nabla f_i(x) \right\Vert^2 \right] \leq \sigma^2, \forall x
\numberthis \label{eqn9}
\end{align*}

The gradient of all $f_i$'s is bounded:
\begin{align*}
\E_{i}\left[ \left\Vert \nabla f_i(x) - \nabla f(x) \right\Vert^2 \right] \leq s^2, \forall x
\numberthis \label{eqn10}
\end{align*}

\item[4.] \textbf{Bound for Parameters during Averaging:} (Based on Assumption 3.2 of \cite{ryabinin2021moshpit})
The average difference between the $i^{th}$ cluster's parameters and the global average of the parameters at the time of averaging is bounded i.e., for any integer $a \geq 0$ there exists a value $ \Delta_a \geq 0$ such that
\begin{align*}
\E\left[ \dfrac{1}{C} \sum_{i=1}^{C} \left\Vert x_i^{a\kappa} - x^{a\kappa} \right\Vert^2 \right] \leq \eta^2 \Delta_a^2
\numberthis \label{eqn11}
\end{align*}

where $\kappa$ is the communication period and $x^t = \dfrac{1}{C} \sum_{i=1}^{C} x_i^t$.

\item[5.] \textbf{Bound for Delay Values:}
All staleness values are bounded i.e. 
$\max_t \tau_t \leq T$

\item[6.] \textbf{Independence:} The random variables in $\left[ b_{i,t} \right]_{t = 0,1,2,...}$ and $\left[i_t\right]_{t = 0,1,2,...}$ are independent of each other.

\end{assumption}

Under the aforementioned assumptions we detail the following theorem and corollary that describe the convergence rate of the defined optimisation problem, the full derivations of which are shown in Appendix \ref{appendixmaintheoremproof} and \ref{appendixcorollaryproof}.

\subsubsection{Notations and Definitions Used}
\label{notations}

We define the following constants for concision:
\begin{align*}
    A_1 =& 1-\dfrac{24\eta^2L^2e(\kappa-1)^2}{C^2N_m}\\
    A_2 =& \dfrac{\eta^2L}{C^2N_m} - \dfrac{\eta}{2CN_m} + \dfrac{2\eta^3L^2T^2}{C^3N_{m}^{2}} + \dfrac{4\eta^2e(\kappa-1)^2}{A_1C^2N_m}\left[\dfrac{6\eta^2L^3}{C^2N_m}+\dfrac{\eta L^2}{CN_m} + \dfrac{12\eta^3L^4T^2}{C^3N_{m}^{2}}\right]\\ 
    A_3 =& \dfrac{1}{A_1}\left(\dfrac{12\eta^2L^3}{C^2N_m} + \dfrac{\eta L^2}{CN_m}\right)\left(\Delta_{a}^{2} + \dfrac{8es^2(\kappa-1)^2}{CN_m} + \dfrac{e\sigma^2(\kappa-1)}{CN_m}\right) 
\end{align*}

\begin{theorem}
\label{theorem1}
Under Assumptions 1-6 and while $\eta$ satisfies the constraints set by $A_1 > 0$, $A_2 \leq 0$, $A_3 \leq 1$ and $\frac{2\eta LT^2}{CN_m}\leq 1$, the proposed algorithm satisfies:
\begin{align*}
\dfrac{1}{K} \sum_{t=0}^{K-1} \E\left[ \left\Vert \nabla f(x^t) \right\Vert^2 \right] \leq \dfrac{2CN_m}{\eta K}(f(x^0) - f(x^*)) + \dfrac{2\eta}{C}(2N_m + L(\sigma^2 + 8s^2))
\numberthis \label{eqn12}
\end{align*}
\end{theorem}

\begin{corollary}
\label{corollary1.1}
In accordance with Theorem \ref{theorem1}, by setting
\begin{align*}
\eta = \dfrac{C\sqrt{N_m}}{\sqrt{K\left[ 2N_m + L(\sigma^2 + 8s^2) \right]}}
\numberthis \label{eqn13}
\end{align*}

and making the number of iterations, $K$, large enough
\begin{align*}
K \geq \dfrac{4L^2}{2N_m + L(\sigma^2 + 8s^2)} \max
\left\{
\begin{aligned}
&5T^4, 25N_m, 20e(\kappa -1)^2 N_m^{\frac{1}{3}}, \\ &9T\sqrt{e}(\kappa-1),  24L\Delta_a^2C, \\ &\dfrac{24Le(\kappa-1)(\sigma^2+8s^2(\kappa-1))}{N_m},  \dfrac{16L^2\Delta_a^2C^2}{N_m}, \\ &\dfrac{16L^2e^2(\kappa-1)^2(\sigma^2+8s^2(\kappa-1))^2}{N_m^3} 
\end{aligned}
\right\}
\numberthis \label{eqn14}
\end{align*}

we get the following convergence rate:
\begin{align*}
\dfrac{1}{K} \sum_{t=0}^{K-1} \E\left[ \left\Vert \nabla f(x^t) \right\Vert^2 \right] \leq \dfrac{2(f(x^0) - f(x^*))}{\sqrt{K}} \sqrt{ 2N_m^2 + LN_m(\sigma^2 + 8s^2)}
\numberthis \label{eqn15}
\end{align*}
\end{corollary}

It can be observed that for values of $K$ that are large enough, the convergence rate of the proposed algorithm is $O\left(\frac{1}{\sqrt{K}}\right)$ which is consistent with the convergence rate of Local-SGD for convex and non-convex optimizations \cite{li2019communication, dekel2012optimal, ghadimi2013stochastic}. 

We can see that if the staleness parameter $T$ is bounded by $O\left(K^{\frac{1}{4}}\right)$, linear speedup is achieved. Note that in case $\kappa - 1 > T^3$, the bound for staleness is $O\left(K\right)$ allowing for convergence with the same iteration complexity for larger number of workers and hence larger magnitudes of staleness compared to 
\cite{lian2015asynchronous}.

In the proposed system, each update on any of the clusters is considered an increment in time-step. Consequently the greater the number of clusters, the greater is the number of update iterations that can be achieved within the same time period allowing for a linear speed up with respect to the number of participating clusters. Thus, we can even allow for clusters to join dynamically in the global averaging graph after downloading the latest global average of model parameters and participate in subsequent averaging rounds leading to an increase in convergence efficiency.

\subsection{Implementation Details}

\subsubsection{Parallel Communication and Training}

On every peer we maintain separate buffers for forward and backward passes that are responsible for receiving and storing the activations and gradients from the neighbouring peers in the cluster's computation graph. The communication thread responsible for populating the buffers and sending data to other peers is run on a separate process parallel to the training process. The training process is responsible for conducting forward and backward passes along with gradient updates as and when the respective buffers are populated. To avoid race conditions during additions or deletions in the buffers, both processes use \textit{Blocking Locks} to prevent simultaneous modifications. Once computations are done by the training process, the communication process sends the required data to the appropriate peers only when the respective buffer of the receiving worker is empty, i.e. the training process on that peer has popped values from it's buffer for computation.

\subsubsection{Cluster Formation Methodology}
\label{clusterformation}

The peak memory usage of a model constitutes of the forward/backward pass size and memory taken up by the parameters. The forward/backward size can serve as a good approximation for the memory required to process one batch of data. Let the peak memory usage of the entire model (which comprises of the above components) be: 
\begin{align*}
M = size(\text{data batch}) \times size(\text{forward/backward pass}) + size(\text{parameters})
\numberthis \label{eqn16}
\end{align*}

We define a pool of tuples containing usable RAM and bandwidth values for $N$ nodes: 
\begin{align*}
P = \left[(r_1,b_1),(r_2,b_2),...,(r_i, b_i)\right] \quad \forall i\in \left[1,2,...,N\right]
\end{align*}

where $r_i$ and $b_i$ represents the RAM and steady bandwidth of the $i^{\text{th}}$ compute node. Let's say we form $Q$ optimal clusters.
Our aim is to find a new set $C = \left[c_i\right]$ where $c_i$ can take discrete values in $\left[1,2,...,Q\right]$. In other words, every $(r_i,b_i)$ is mapped to a corresponding discrete valued $c_i$ which represents the ID of the cluster that $(r_i,b_i)$ belongs to. 
Note that $P$ and $C$ have the same number of elements. Since we need to effectively fit the entire model of size $M$ on each cluster, the first constraint in our optimization problem is that the sum of all peer's RAM values need to be greater than $M$. Across all clusters in a training session, the sum of individual times to transfer data taken by each node in a cluster should be comparable. This is equivalent to saying that the maximum difference in sums of data transfer times across any 2 clusters needs to be minimized.This forms our second constraint. 

This is an NP-hard problem(meaning there's no polynomial time solution), a heuristic/greedy approach might be better suited for finding optimal solutions. This optimization problem can be solved using a Meta-Heuristic Genetic Algorithm \cite{goyal2020genetic}. 
We initialize a random population of individual chromosomes, iteratively evolve them across generations while introducing mutations and select the best performing ones based on a fitness function to create new successor generations \cite{katoch2021review}. 

The aforementioned set $C$ serves as a representation for an individual chromosome, which are assigned random values initially. A collection of such chromosomes are spawned to form a population, which are together evolved across generations. A penalty based fitness function (that uses the problem constraints) is then used to evaluate all individuals in a population out of which, the best ones are selected for crossover, resulting in offsprings that make up the next generation's population. The individual with the best fitness value observed is tracked across all generations and returned. Since each cluster formed will typically contain a mix of nodes with varying RAM values, the main model parameters will undergo a proportionate non-uniform split such that each peer can accomodate the largest chunk possible. 

For a communication infrastructure orchestrated over the internet, fault tolerance plays a crucial role in ensuring successful model training.  Ravnest allows new peers to join any ongoing distributed training sessions. Based on the session's fault tolerance requirements, they can either join a pre-existing cluster and participate as backups nodes or wait till enough new peers are available to form an entirely new cluster that, once formed, will fetch the latest set of aggregated parameters and join the ring topology. The latter case provides an advantage, since having more clusters improves the efficiency of the training process as was shown in Corollary \ref{corollary1.1}. In the former scenario, the new peer will be mapped as a backup for the most unreliable node in the cluster. An extra pair of communication channels will be spawned that shares a copy of forward/backward pass results with the backup node so that it can take over in case the primary node disconnects.

\section{Experiments}
\label{experiments}

This section elucidates upon a series of experimental training sessions conducted to evaluate the performance and efficiency of our proposed methodology across both single and multi-cluster setups with comparisons to baseline synchronous training.

\subsection{Single Cluster Experiments with Homogeneous Nodes}

We trained a ResNet-50 network \cite{he2015deep} on the Tiny ImageNet dataset \cite{wu2017tiny} for 50 epochs with a batch size of 100. SGD was utilized with a momentum of 0.9, weight decay of 0.0005 and learning rate of 0.01 which was reduced by a factor of 10 every 30 epochs. Images were pre-processed by scaling to 224x224x3, then normalized to [0,1], with a mean of [0.485, 0.456, 0.406] and standard deviation of [0.229, 0.224, 0.225]. We also trained an Inception-V3 model \cite{szegedy2015rethinking} on the CIFAR-10 dataset \cite{krizhevsky2009learning} for 50 epochs with a batch size of 64 and SGD with a momentum of 0.9, weight decay of 0.0005 and learning rate of 0.01, reducing by a factor of 10 every 30 epochs. Images were pre-processed by normalization (mean: [0.4914, 0.4822, 0.4465], standard deviation: [0.2023, 0.1994, 0.2010]). Augmentation techniques including random flipping, cropping and padding were applied for both datasets. While these settings provide a robust starting point, it is important to note that further fine-tuning of hyperparameters could potentially lead to improved performance.

\begin{figure}[htp]
  \centering
  \subfloat[ResNet-50 model trained on Tiny ImageNet]{\includegraphics[width=0.45\textwidth]{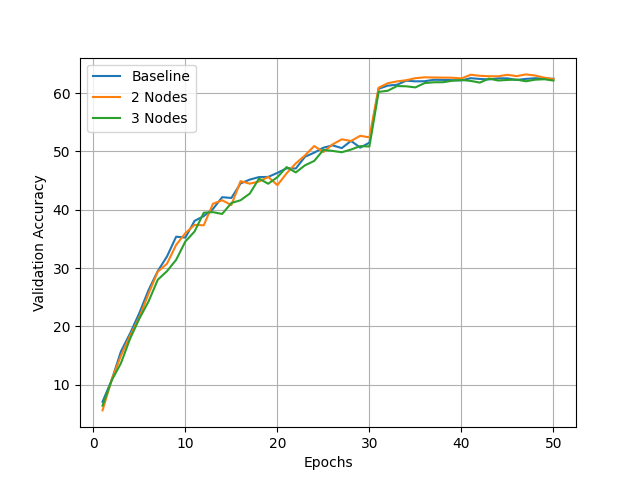}\label{fig:resnet_val_accuracies}}
  \hfill
  \subfloat[Inception-V3 model trained on CIFAR-10]{\includegraphics[width=0.45\textwidth]{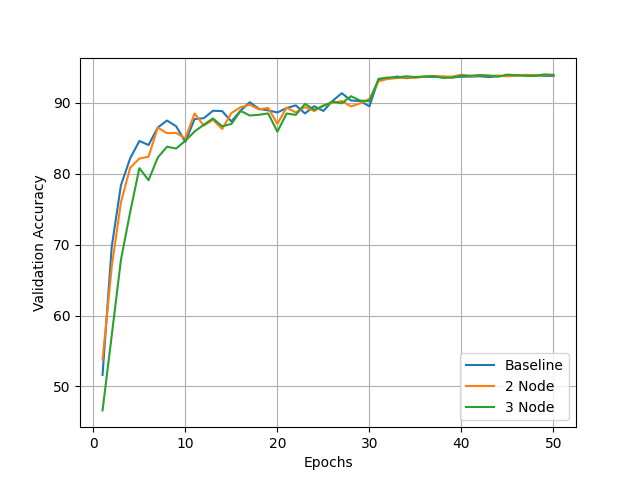}\label{fig:inception_val_accuracies}}
  \caption{Comparison of validation accuracies over epochs}.
\end{figure}

For both models, conventional synchronous training was performed as a baseline, followed by decentralized training using proposed approach across 2 and 3 homogeneous nodes (within a single cluster), all hosted on Nvidia A10G GPUs with 24 GB VRAM. From Figures \ref{fig:resnet_val_accuracies} and \ref{fig:inception_val_accuracies}, it is evident that as the number of nodes within the single cluster grows, initial validation accuracy registers lower values compared to scenarios with fewer nodes. This can be ascribed to the accumulation of more stagnant gradients as the size of the cluster increases. However, as training advances, the rates of convergence are at par with those of the baselines. We also observed that the average memory usage per node reduced by around 56\% (3 nodes case) and by 42\% (2 nodes case) for both models when compared to conventional training on a single GPU instance with the same batch sizes.

\subsection{Multi-Cluster Experiments with Heterogeneous Nodes}
We evaluated a multi-cluster setup using the proposed method on the BERT model following the BERT base (uncased) architecture \cite{devlin2018bert}. Pretraining was done on Masked LM (MLM) and Next Sentence Prediction (NSP) tasks using the LAMB optimizer \cite{You2020Large} with a learning rate of 0.00176 and weight decay of 0.01, on the WikiText-103 dataset \cite{merity2016pointer}. A linear scheduler with 5000 warmup steps was also utilized. The sum of the losses for these 2 tasks was used as the convergence metric. The multi-cluster setup consisted of 10 nodes across A10G, V100 and T4 GPUs hosted on cloud instances. They were grouped into 4 clusters by the cluster formation genetic algorithm in order to ensure that the total compute times were similar across all clusters. 
\begin{figure}[htp]
    \centering    \includegraphics[width=0.45\textwidth]{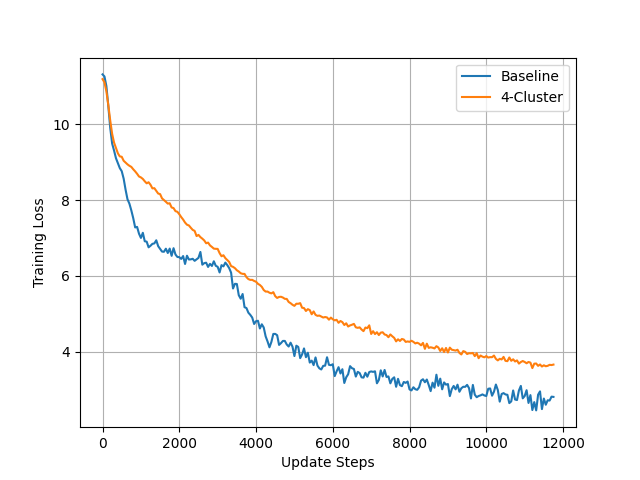}
    \caption{BERT pre-training on multi-cluster setup} \label{fig:bertloss}
\end{figure}

Each cluster consisted of either 2 or 3 nodes, allowing for varying degrees of staleness, and gradients were accumulated over 32 batches of 8 samples before performing updates. Parameter averaging was done whenever each cluster completed 1400 update iterations. Conventional synchronous baseline training was done with the same hyperparameters.  Figure \ref{fig:bertloss} shows that in 11797 update iterations, the proposed algorithm achieves a learning curve comparable to that of the baseline. Training for more iterations or increasing the number of clusters can lead to even better results. Additionally, we noted that the average GPU memory usage across all nodes was approximately 67\% less than that seen during training on a single GPU instance with the same batch size of 8.

\section{Conclusion}
This research paper proposes an novel asynchronous parallel decentralised training paradigm that leverages the strengths of both data and model parallelism techniques. It incorporates \textit{Parallel Multi-Ring All-Reduce} for efficient parameter averaging across clusters,  to train contemporary deep learning models across heterogeneous consumer grade PCs. \textit{Zero-bubble Asynchronous Model Parallelism} within each cluster further minimizes the idle time bubble on every node. Comprehensive theoretical analysis was performed to validate the convergence and linear speedup characteristics of the approach.

\bibliographystyle{plain}
\bibliography{pub}

\begin{thebibliography}{10}

\bibitem{alayrac2022flamingo}
Jean-Baptiste Alayrac, Jeff Donahue, Pauline Luc, Antoine Miech, Iain Barr,
  Yana Hasson, Karel Lenc, Arthur Mensch, Katherine Millican, Malcolm Reynolds,
  et~al.
\newblock Flamingo: a visual language model for few-shot learning.
\newblock {\em Advances in Neural Information Processing Systems},
  35:23716--23736, 2022.

\bibitem{brown2020language}
Tom Brown, Benjamin Mann, Nick Ryder, Melanie Subbiah, Jared~D Kaplan, Prafulla
  Dhariwal, Arvind Neelakantan, Pranav Shyam, Girish Sastry, Amanda Askell,
  et~al.
\newblock Language models are few-shot learners.
\newblock {\em Advances in neural information processing systems},
  33:1877--1901, 2020.

\bibitem{chang2023survey}
Yupeng Chang, Xu~Wang, Jindong Wang, Yuan Wu, Linyi Yang, Kaijie Zhu, Hao Chen,
  Xiaoyuan Yi, Cunxiang Wang, Yidong Wang, Wei Ye, Yue Zhang, Yi~Chang,
  Philip~S. Yu, Qiang Yang, and Xing Xie.
\newblock A survey on evaluation of large language models, 2023.

\bibitem{cipar2013solving}
James Cipar, Qirong Ho, Jin~Kyu Kim, Seunghak Lee, Gregory~R Ganger, Garth
  Gibson, Kimberly Keeton, and Eric Xing.
\newblock Solving the straggler problem with bounded staleness.
\newblock In {\em 14th Workshop on Hot Topics in Operating Systems (HotOS
  XIV)}, 2013.

\bibitem{croitoru2023diffusion}
Florinel-Alin Croitoru, Vlad Hondru, Radu~Tudor Ionescu, and Mubarak Shah.
\newblock Diffusion models in vision: A survey.
\newblock {\em IEEE Transactions on Pattern Analysis and Machine Intelligence},
  2023.

\bibitem{cui2021asynchronous}
Xiaodong Cui, Wei Zhang, Abdullah Kayi, Mingrui Liu, Ulrich Finkler, Brian
  Kingsbury, George Saon, and David Kung.
\newblock Asynchronous decentralized distributed training of acoustic models.
\newblock {\em IEEE/ACM Transactions on Audio, Speech, and Language
  Processing}, 29:3565--3576, 2021.

\bibitem{dai2023efficient}
Fei Dai, Yawen Chen, Zhiyi Huang, Haibo Zhang, and Fangfang Zhang.
\newblock Efficient all-reduce for distributed dnn training in optical
  interconnect systems.
\newblock In {\em Proceedings of the 28th ACM SIGPLAN Annual Symposium on
  Principles and Practice of Parallel Programming}, pages 422--424, 2023.

\bibitem{dekel2012optimal}
Ofer Dekel, Ran Gilad-Bachrach, Ohad Shamir, and Lin Xiao.
\newblock Optimal distributed online prediction using mini-batches.
\newblock {\em Journal of Machine Learning Research}, 13(1), 2012.

\bibitem{devlin2018bert}
Jacob Devlin, Ming-Wei Chang, Kenton Lee, and Kristina Toutanova.
\newblock Bert: Pre-training of deep bidirectional transformers for language
  understanding.
\newblock {\em arXiv preprint arXiv:1810.04805}, 2018.

\bibitem{diskin2021distributed}
Michael Diskin, Alexey Bukhtiyarov, Max Ryabinin, Lucile Saulnier, Anton
  Sinitsin, Dmitry Popov, Dmitry~V Pyrkin, Maxim Kashirin, Alexander Borzunov,
  Albert Villanova~del Moral, et~al.
\newblock Distributed deep learning in open collaborations.
\newblock {\em Advances in Neural Information Processing Systems},
  34:7879--7897, 2021.

\bibitem{dutta2021slow}
Sanghamitra Dutta, Jianyu Wang, and Gauri Joshi.
\newblock Slow and stale gradients can win the race.
\newblock {\em IEEE Journal on Selected Areas in Information Theory},
  2(3):1012--1024, 2021.

\bibitem{elgabli2020communication}
Anis Elgabli, Jihong Park, Amrit~S Bedi, Mehdi Bennis, and Vaneet Aggarwal.
\newblock Communication efficient framework for decentralized machine learning.
\newblock In {\em 2020 54th Annual Conference on Information Sciences and
  Systems (CISS)}, pages 1--5. IEEE, 2020.

\bibitem{ghadimi2013stochastic}
Saeed Ghadimi and Guanghui Lan.
\newblock Stochastic first-and zeroth-order methods for nonconvex stochastic
  programming.
\newblock {\em SIAM Journal on Optimization}, 23(4):2341--2368, 2013.

\bibitem{goyal2020genetic}
Gunjan Goyal, Pankaj~Kumar Srivastava, and Dinesh~CS Bisht.
\newblock Genetic algorithm: a metaheuristic approach of optimization.
\newblock {\em cf there}, pages 27--4, 2020.

\bibitem{goyal2017accurate}
Priya Goyal, Piotr Doll{\'a}r, Ross Girshick, Pieter Noordhuis, Lukasz
  Wesolowski, Aapo Kyrola, Andrew Tulloch, Yangqing Jia, and Kaiming He.
\newblock Accurate, large minibatch sgd: Training imagenet in 1 hour.
\newblock {\em arXiv preprint arXiv:1706.02677}, 2017.

\bibitem{guan2019xpipe}
Lei Guan, Wotao Yin, Dongsheng Li, and Xicheng Lu.
\newblock Xpipe: Efficient pipeline model parallelism for multi-gpu dnn
  training.
\newblock {\em arXiv preprint arXiv:1911.04610}, 2019.

\bibitem{harlap2018pipedream}
Aaron Harlap, Deepak Narayanan, Amar Phanishayee, Vivek Seshadri, Nikhil
  Devanur, Greg Ganger, and Phil Gibbons.
\newblock Pipedream: Fast and efficient pipeline parallel dnn training.
\newblock {\em arXiv preprint arXiv:1806.03377}, 2018.

\bibitem{he2015deep}
Kaiming He, Xiangyu Zhang, Shaoqing Ren, and Jian Sun.
\newblock Deep residual learning for image recognition, 2015.

\bibitem{huang2019gpipe}
Yanping Huang, Youlong Cheng, Ankur Bapna, Orhan Firat, Dehao Chen, Mia Chen,
  HyoukJoong Lee, Jiquan Ngiam, Quoc~V Le, Yonghui Wu, et~al.
\newblock Gpipe: Efficient training of giant neural networks using pipeline
  parallelism.
\newblock {\em Advances in neural information processing systems}, 32, 2019.

\bibitem{katoch2021review}
Sourabh Katoch, Sumit~Singh Chauhan, and Vijay Kumar.
\newblock A review on genetic algorithm: past, present, and future.
\newblock {\em Multimedia tools and applications}, 80:8091--8126, 2021.

\bibitem{krizhevsky2009learning}
Alex Krizhevsky, Geoffrey Hinton, et~al.
\newblock Learning multiple layers of features from tiny images.
\newblock {\em Technical report}, 2009.

\bibitem{li2014scaling}
Mu~Li, David~G Andersen, Jun~Woo Park, Alexander~J Smola, Amr Ahmed, Vanja
  Josifovski, James Long, Eugene~J Shekita, and Bor-Yiing Su.
\newblock Scaling distributed machine learning with the parameter server.
\newblock In {\em 11th USENIX Symposium on operating systems design and
  implementation (OSDI 14)}, pages 583--598, 2014.

\bibitem{li2020pytorch}
Shen Li, Yanli Zhao, Rohan Varma, Omkar Salpekar, Pieter Noordhuis, Teng Li,
  Adam Paszke, Jeff Smith, Brian Vaughan, Pritam Damania, et~al.
\newblock Pytorch distributed: Experiences on accelerating data parallel
  training.
\newblock {\em arXiv preprint arXiv:2006.15704}, 2020.

\bibitem{li2019communication}
Xiang Li, Wenhao Yang, Shusen Wang, and Zhihua Zhang.
\newblock Communication efficient decentralized training with multiple local
  updates.
\newblock {\em arXiv preprint arXiv:1910.09126}, 5:6, 2019.

\bibitem{lian2015asynchronous}
Xiangru Lian, Yijun Huang, Yuncheng Li, and Ji~Liu.
\newblock Asynchronous parallel stochastic gradient for nonconvex optimization.
\newblock {\em Advances in neural information processing systems}, 28, 2015.

\bibitem{mattson2020mlperf}
Peter Mattson, Christine Cheng, Gregory Diamos, Cody Coleman, Paulius
  Micikevicius, David Patterson, Hanlin Tang, Gu-Yeon Wei, Peter Bailis, Victor
  Bittorf, et~al.
\newblock Mlperf training benchmark.
\newblock {\em Proceedings of Machine Learning and Systems}, 2:336--349, 2020.

\bibitem{merity2016pointer}
Stephen Merity, Caiming Xiong, James Bradbury, and Richard Socher.
\newblock Pointer sentinel mixture models.
\newblock {\em arXiv preprint arXiv:1609.07843}, 2016.

\bibitem{narayanan2021memory}
Deepak Narayanan, Amar Phanishayee, Kaiyu Shi, Xie Chen, and Matei Zaharia.
\newblock Memory-efficient pipeline-parallel dnn training.
\newblock In {\em International Conference on Machine Learning}, pages
  7937--7947. PMLR, 2021.

\bibitem{nvidiaNVIDIACollective}
{N}{V}{I}{D}{I}{A}.
\newblock {C}ollective {C}ommunications {L}ibrary ({N}{C}{C}{L}).
\newblock \url{https://developer.nvidia.com/nccl}, 2019.

\bibitem{leelazero}
{Pascutto, Gian-Carlo}.
\newblock Leela zero, 2019.

\bibitem{patarasuk2009bandwidth}
Pitch Patarasuk and Xin Yuan.
\newblock Bandwidth optimal all-reduce algorithms for clusters of workstations.
\newblock {\em Journal of Parallel and Distributed Computing}, 69(2):117--124,
  2009.

\bibitem{rabenseifner2004optimization}
Rolf Rabenseifner.
\newblock Optimization of collective reduction operations.
\newblock In {\em Computational Science-ICCS 2004: 4th International
  Conference, Krak{\'o}w, Poland, June 6-9, 2004, Proceedings, Part I 4}, pages
  1--9. Springer, 2004.

\bibitem{recht2011hogwild}
Benjamin Recht, Christopher Re, Stephen Wright, and Feng Niu.
\newblock Hogwild!: A lock-free approach to parallelizing stochastic gradient
  descent.
\newblock {\em Advances in neural information processing systems}, 24, 2011.

\bibitem{ryabinin2021moshpit}
Max Ryabinin, Eduard Gorbunov, Vsevolod Plokhotnyuk, and Gennady Pekhimenko.
\newblock Moshpit sgd: Communication-efficient decentralized training on
  heterogeneous unreliable devices.
\newblock {\em Advances in Neural Information Processing Systems},
  34:18195--18211, 2021.

\bibitem{ryabinin2020towards}
Max Ryabinin and Anton Gusev.
\newblock Towards crowdsourced training of large neural networks using
  decentralized mixture-of-experts.
\newblock {\em Advances in Neural Information Processing Systems},
  33:3659--3672, 2020.

\bibitem{scaman2019optimal}
Kevin Scaman, Francis Bach, S{\'e}bastien Bubeck, Yin~Tat Lee, and Laurent
  Massouli{\'e}.
\newblock Optimal convergence rates for convex distributed optimization in
  networks.
\newblock {\em Journal of Machine Learning Research}, 20:1--31, 2019.

\bibitem{sergeev2018horovod}
Alexander Sergeev and Mike Del~Balso.
\newblock Horovod: fast and easy distributed deep learning in tensorflow.
\newblock {\em arXiv preprint arXiv:1802.05799}, 2018.

\bibitem{shallue2019measuring}
Christopher~J Shallue, Jaehoon Lee, Joseph Antognini, Jascha Sohl-Dickstein,
  Roy Frostig, and George~E Dahl.
\newblock Measuring the effects of data parallelism on neural network training.
\newblock {\em Journal of Machine Learning Research}, 20(112):1--49, 2019.

\bibitem{shoeybi2019megatron}
Mohammad Shoeybi, Mostofa Patwary, Raul Puri, Patrick LeGresley, Jared Casper,
  and Bryan Catanzaro.
\newblock Megatron-lm: Training multi-billion parameter language models using
  model parallelism.
\newblock {\em arXiv preprint arXiv:1909.08053}, 2019.

\bibitem{szegedy2015rethinking}
Christian Szegedy, Vincent Vanhoucke, Sergey Ioffe, Jonathon Shlens, and
  Zbigniew Wojna.
\newblock Rethinking the inception architecture for computer vision, 2015.

\bibitem{tanaka2018large}
Kenji Tanaka, Yuki Arikawa, Kenji Kawai, Junichi Kato, Tsuyoshi Ito, Huy~Cu
  Ngo, Kazutaka Morita, Fumiaki Miura, Takeshi Sakamoto, and Satoshi
  Shigematsu.
\newblock Large-message size allreduce at wire speed for distributed deep
  learning.
\newblock In {\em Poster session presented at SC18, and Analysis}, 2018.

\bibitem{ueno2019exhaustive}
Yuichiro Ueno and Rio Yokota.
\newblock Exhaustive study of hierarchical allreduce patterns for large
  messages between gpus.
\newblock In {\em 2019 19th IEEE/ACM International Symposium on Cluster, Cloud
  and Grid Computing (CCGRID)}, pages 430--439. IEEE, 2019.

\bibitem{uribe2020dual}
C{\'e}sar~A Uribe, Soomin Lee, Alexander Gasnikov, and Angelia Nedi{\'c}.
\newblock A dual approach for optimal algorithms in distributed optimization
  over networks.
\newblock In {\em 2020 Information Theory and Applications Workshop (ITA)},
  pages 1--37. IEEE, 2020.

\bibitem{valiant1990bridging}
Leslie~G Valiant.
\newblock A bridging model for parallel computation.
\newblock {\em Communications of the ACM}, 33(8):103--111, 1990.

\bibitem{villalobos2022machine}
Pablo Villalobos, Jaime Sevilla, Tamay Besiroglu, Lennart Heim, Anson Ho, and
  Marius Hobbhahn.
\newblock Machine learning model sizes and the parameter gap.
\newblock {\em arXiv preprint arXiv:2207.02852}, 2022.

\bibitem{wei2022emergent}
Jason Wei, Yi~Tay, Rishi Bommasani, Colin Raffel, Barret Zoph, Sebastian
  Borgeaud, Dani Yogatama, Maarten Bosma, Denny Zhou, Donald Metzler, et~al.
\newblock Emergent abilities of large language models.
\newblock {\em arXiv preprint arXiv:2206.07682}, 2022.

\bibitem{wei2021deploying}
Jia Wei, Xingjun Zhang, Zeyu Ji, Jingbo Li, and Zheng Wei.
\newblock Deploying and scaling distributed parallel deep neural networks on
  the tianhe-3 prototype system.
\newblock {\em Scientific Reports}, 11(1):20244, 2021.

\bibitem{wu2017tiny}
Jiayu Wu, Qixiang Zhang, and Guoxi Xu.
\newblock Tiny imagenet challenge.
\newblock {\em Technical report}, 2017.

\bibitem{wu2023next}
Shengqiong Wu, Hao Fei, Leigang Qu, Wei Ji, and Tat-Seng Chua.
\newblock Next-gpt: Any-to-any multimodal llm.
\newblock {\em arXiv preprint arXiv:2309.05519}, 2023.

\bibitem{xu2020acceleration}
An~Xu, Zhouyuan Huo, and Heng Huang.
\newblock On the acceleration of deep learning model parallelism with
  staleness.
\newblock In {\em Proceedings of the IEEE/CVF Conference on Computer Vision and
  Pattern Recognition}, pages 2088--2097, 2020.

\bibitem{xu2021dp}
Jie Xu, Wei Zhang, and Fei Wang.
\newblock A $(\text{DP})^ 2$ sgd: Asynchronous decentralized parallel
  stochastic gradient descent with differential privacy.
\newblock {\em IEEE transactions on pattern analysis and machine intelligence},
  44(11):8036--8047, 2021.

\bibitem{yang2021pipemare}
Bowen Yang, Jian Zhang, Jonathan Li, Christopher R{\'e}, Christopher Aberger,
  and Christopher De~Sa.
\newblock Pipemare: Asynchronous pipeline parallel dnn training.
\newblock {\em Proceedings of Machine Learning and Systems}, 3:269--296, 2021.

\bibitem{yang2023diffusion}
Ling Yang, Zhilong Zhang, Yang Song, Shenda Hong, Runsheng Xu, Yue Zhao, Wentao
  Zhang, Bin Cui, and Ming-Hsuan Yang.
\newblock Diffusion models: A comprehensive survey of methods and applications.
\newblock {\em ACM Computing Surveys}, 56(4):1--39, 2023.

\bibitem{You2020Large}
Yang You, Jing Li, Sashank Reddi, Jonathan Hseu, Sanjiv Kumar, Srinadh
  Bhojanapalli, Xiaodan Song, James Demmel, Kurt Keutzer, and Cho-Jui Hsieh.
\newblock Large batch optimization for deep learning: Training bert in 76
  minutes.
\newblock In {\em International Conference on Learning Representations}, 2020.

\bibitem{zhang2015staleness}
Wei Zhang, Suyog Gupta, Xiangru Lian, and Ji~Liu.
\newblock Staleness-aware async-sgd for distributed deep learning.
\newblock {\em arXiv preprint arXiv:1511.05950}, 2015.

\bibitem{zhao2020distributed}
Weijie Zhao, Deping Xie, Ronglai Jia, Yulei Qian, Ruiquan Ding, Mingming Sun,
  and Ping Li.
\newblock Distributed hierarchical gpu parameter server for massive scale deep
  learning ads systems.
\newblock {\em Proceedings of Machine Learning and Systems}, 2:412--428, 2020.

\bibitem{zhao2023pytorch}
Yanli Zhao, Andrew Gu, Rohan Varma, Liang Luo, Chien-Chin Huang, Min Xu, Less
  Wright, Hamid Shojanazeri, Myle Ott, Sam Shleifer, et~al.
\newblock Pytorch fsdp: experiences on scaling fully sharded data parallel.
\newblock {\em arXiv preprint arXiv:2304.11277}, 2023.

\bibitem{zheng2017asynchronous}
Shuxin Zheng, Qi~Meng, Taifeng Wang, Wei Chen, Nenghai Yu, Zhi-Ming Ma, and
  Tie-Yan Liu.
\newblock Asynchronous stochastic gradient descent with delay compensation.
\newblock In {\em International Conference on Machine Learning}, pages
  4120--4129. PMLR, 2017.

\bibitem{zhuang2023optimizing}
Yonghao Zhuang, Hexu Zhao, Lianmin Zheng, Zhuohan Li, Eric Xing, Qirong Ho,
  Joseph Gonzalez, Ion Stoica, and Hao Zhang.
\newblock On optimizing the communication of model parallelism.
\newblock {\em Proceedings of Machine Learning and Systems}, 5, 2023.

\end{thebibliography}

\begin{appendix}
\label{appendix}

\section*{APPENDIX}
\section{Proof for Theorem \ref{theorem1}}
\label{appendixmaintheoremproof}

Main Proof:

We start with the Global-Scale (across clusters) update step equation:

\begin{align*}
    x^{t+1} = x^{t} - \dfrac{\eta}{C}S_{n_{i_t,t}} \nabla F(x_{i_t}^{t-\tau_t}, b_{i_t}^{t-\tau_t})
\numberthis \label{eqn17}
\end{align*}

From the Lipschitzian gradient assumption \ref{Assumptions}:2, we have:

\begin{align*}
    f(x^{t+1}) \leq f(x^{t}) - \dfrac{\eta}{C} \left\langle \nabla f(x^{t}), S_{n_{i_t,t}} g_{i_t}^{t-\tau_t} \right\rangle + \dfrac{\eta^{2}L}{2} \left\Vert \dfrac{S_{n_{i_t,t}} g_{i_t}^{t-\tau_t}}{C} \right\Vert^2
\numberthis \label{eqn18}
\end{align*}

Taking full expectation on both sides:

\begin{align*}
    \E\left[ f(x^{t+1}) \right] \leq &\E\left[ f(x^{t}) \right] - \dfrac{\eta}{C}\E\left\langle \nabla f(x^{t}), \dfrac{1}{C}\sum_{i_t=1}^{C} \dfrac{\nabla f_{i_t}(x_{i_t}^{t-\tau_t})}{N_{i_t}} \right\rangle\\ &+ \dfrac{\eta^2L}{2C^2}\E\left[ \dfrac{1}{C}\sum_{i_t=1}^{C} \dfrac{\left\Vert g_{i_t}^{t-\tau_t} \right\Vert^2}{N_{i_t}} \right]\\ \implies \E\left[ f(x^{t+1}) \right] \leq &\E\left[ f(x^{t}) \right] - \dfrac{\eta}{CN_m}\E\left\langle \nabla f(x^{t}), \dfrac{1}{C}\sum_{i_t=1}^{C} \nabla f_{i_t}(x_{i_t}^{t-\tau_t}) \right\rangle\\ &+ \dfrac{\eta^2L}{2C^2N_m}\E\left[ \dfrac{1}{C}\sum_{i_t=1}^{C} \left\Vert g_{i_t}^{t-\tau_t} \right\Vert^2 \right]
\end{align*}

Using the property: $\left\Vert a + b \right\Vert^2 \leq 2\left\Vert a \right\Vert^2 + 2\left\Vert b \right\Vert^2$

\begin{align*}
    \E\left[ f(x^{t+1}) \right] \leq &\E\left[ f(x^{t}) \right] - \dfrac{\eta}{2CN_m}\E\left[ \left\Vert \nabla f(x^t) \right\Vert^2 \right]\\ &- \dfrac{\eta}{2CN_m}\E\left[\left\Vert \dfrac{1}{C}\sum_{i_t=1}^{C} \nabla f_{i_t}(x_{i_t}^{t-\tau_t}) \right\Vert^2\right]\\ &+ \dfrac{\eta}{2CN_m}\underbrace{\E\left[ \left\Vert \nabla f(x^t) - \dfrac{1}{C}\sum_{i_t=1}^{C}\nabla f_{i_t}(x_{i_t}^{t-\tau_t}) \right\Vert^2 \right]}_{T_2}\\ &+ \dfrac{\eta^2L}{2C^2N_m}\underbrace{\E\left[ \dfrac{1}{C}\sum_{i_t=1}^{C} \left\Vert g_{i_t}^{t-\tau_t} \right\Vert^2 \right]}_{T_1}
\numberthis \label{eqn19}
\end{align*}

Expanding the term $T_1$:

\begin{align*}
    T_1 =& \E\left[ \dfrac{1}{C}\sum_{i_t=1}^{C} \left\Vert g_{i_t}^{t-\tau_t} \right\Vert^2 \right]\\  =&\E\left[\dfrac{1}{C}\sum_{i_t=1}^{C}\left\Vert g_{i_t}^{t-\tau_t} - \nabla f_{i_t}(x_{i_t}^{t-\tau_t}) \right\Vert^2\right] + \E\left[ \dfrac{1}{C}\sum_{i_t=1}^{C} \left\Vert \nabla f_{i_t}(x_{i_t}^{t-\tau_t}) \right\Vert^2 \right]\\ \leq & \sigma^2 + \E\left[ \dfrac{1}{C}\sum_{i_t=1}^{C} \left\Vert \nabla f_{i_t}(x_{i_t}^{t-\tau_t}) \right\Vert^2 \right]
\numberthis \label{eqn20}
\end{align*}

Expanding the term $T_2$: 

\begin{align*}
    T_2 =& \E\left[ \left\Vert \nabla f(x^t) - \dfrac{1}{C}\sum_{i_t=1}^{C}\nabla f_{i_t}(x_{i_t}^{t-\tau_t}) \right\Vert^2 \right]\\ =&\E\left[ \left\Vert \nabla f(x^t) - \nabla f(x^{t-\tau_t}) + \nabla f(x^{t-\tau_t}) - \dfrac{1}{C}\sum_{i_t=1}^{C}\nabla f_{i_t}(x_{i_t}^{t-\tau_t}) \right\Vert^2 \right]\\ \leq& 2\E\left[ \left\Vert \nabla f(x^t) - \nabla f(x^{t-\tau_t}) \right\Vert^2 \right]\\ &+ 2\E\left[ \left\Vert \dfrac{1}{C}\sum_{i_t=1}^{C} \left(\nabla f_{i_t}(x^{t-\tau_t}) - \nabla f_{i_t}(x_{i_t}^{t-\tau_t}) \right)\right\Vert^2 \right]\\ \leq& 2L^2\E\left[\left\Vert x^t - x^{t-\tau_t} \right\Vert^2\right] + 2L^2\E\left[\underbrace{\dfrac{1}{C}\sum_{i_t=1}^{C}\left\Vert x^{t-\tau_t} - x_{i_t}^{t-\tau_t} \right\Vert^2}_{\delta^{t-\tau_t}}\right]\\ \leq& 2L^2\underbrace{\E\left[\left\Vert x^t - x^{t-\tau_t} \right\Vert^2\right]}_{T_3} + 2L^2\E\left[\delta^{t-\tau_t}\right]
\numberthis \label{eqn21}
\end{align*}

Expanding the term $T_3$:

\begin{align*}
    T_3 =& \E\left[\left\Vert x^t - x^{t-\tau_t} \right\Vert^2\right]\\ \leq& \E\left[\left\Vert \sum_{j=t-\tau_t}^{t-1}\left( x^{j+1} - x^j\right) \right\Vert^2\right]\\ \leq& \E\left[\left\Vert \sum_{j=t-\tau_t}^{t-1} \dfrac{\eta}{C}S_{n_{i_j}}g_{i_j}^{j-\tau_j} \right\Vert^2\right]
    \end{align*}
    \begin{align*}
    T_3 \leq& \dfrac{\eta^2}{N_m}\E\left[\dfrac{1}{C}\sum_{i_j=1}^{C}\left\Vert \dfrac{1}{C}\sum_{j=t-\tau_t}^{t-1}g_{i_j}^{j-\tau_j} \right\Vert^2\right]\\ \leq& \dfrac{\eta^2\tau_t}{C^2N_m}\sum_{j=t-\tau_t}^{t-1}\E\left[ \dfrac{1}{C}\sum_{i_j=1}^{C}\left\Vert g_{i_j}^{j-\tau_j} \right\Vert^2 \right]\\ \leq& \dfrac{\eta^2\tau_t}{C^2N_m}\sum_{j=t-\tau_t}^{t-1}\left(\sigma^2 + \E\left[\dfrac{1}{C}\sum_{i_j=1}^{C}\left\Vert \nabla f_{i_j}(x_{i_j}^{j-\tau_j}) \right\Vert^2\right]\right)\\ \leq& \dfrac{\eta^2\tau_{t}^{2}\sigma^2}{C^2N_m} + \dfrac{\eta^2\tau_t}{C^2N_m}\sum_{j=t-\tau_t}^{t-1}\E\left[\dfrac{1}{C}\sum_{i_j=1}^{C}\left\Vert \nabla f_{i_j}(x_{i_j}^{j-\tau_j}) \right\Vert^2\right]
\numberthis \label{eqn22}
\end{align*}

Using equations \ref{eqn20}, \ref{eqn21} and \ref{eqn22}, equation \ref{eqn19} becomes:

\begin{align*}
    \E\left[ f(x^{t+1}) \right] \leq &\E\left[ f(x^{t}) \right] - \dfrac{\eta}{2CN_m}\E\left[ \left\Vert \nabla f(x^t) \right\Vert^2 \right]\\ &- \dfrac{\eta}{2CN_m}\E\left[\left\Vert \dfrac{1}{C}\sum_{i_t=1}^{C} \nabla f_{i_t}(x_{i_t}^{t-\tau_t}) \right\Vert^2\right]\\ &+ \dfrac{\eta^2L\sigma^2}{2C^2N_m} + \dfrac{\eta^2L}{2C^2N_m}\underbrace{\E\left[\dfrac{1}{C}\sum_{i_t=1}^{C}\left\Vert \nabla f_{i_t}(x_{i_t}^{t-\tau_t}) \right\Vert^2\right]}_{T_4}\\ &+ \dfrac{\eta L^2}{CN_m}\left( \dfrac{\eta^2\tau_{t}^{2}\sigma^2}{C^2N_m} + \dfrac{\eta^2\tau_t}{C^2N_m}\sum_{j=t-\tau_t}^{t-1}\E\left[\dfrac{1}{C}\sum_{i_j=1}^{C}\left\Vert \nabla f_{i_j}(x_{i_j}^{j-\tau_j}) \right\Vert^2\right] \right) + \dfrac{\eta L^2}{CN_m}\E\left[\delta^{t-\tau_t}\right]
\end{align*}

The term $T_4$ can be simplified using lemma \ref{lemma1}:

\begin{align*}
    \E\left[ f(x^{t+1}) \right] \leq &\E\left[ f(x^{t}) \right] - \dfrac{\eta}{2CN_m}\E\left[ \left\Vert \nabla f(x^t) \right\Vert^2 \right]  + \left(\dfrac{\eta^2L}{C^2N_m} - \dfrac{\eta}{2CN_m}\right)\E\left[\left\Vert\dfrac{1}{C}\sum_{i_t=1}^{C} \nabla f_{i_t}(x_{i_t}^{t-\tau_t})\right\Vert^2\right]\\ &+ \dfrac{\eta^2L\sigma^2}{2C^2N_m} + \dfrac{6\eta^2L^3}{C^2N_m}\E\left[\delta^{t-\tau_t}\right] + \dfrac{4\eta^2Ls^2}{C^2N_m} + \dfrac{\eta^3L^2\tau_{t}^2\sigma^2}{C^3N_{m}^{2}}\\ &+ \dfrac{\eta^3L^2\tau_t}{C^3N_{m}^{2}}\sum_{j=t-\tau_t}^{t-1}\E\left[\dfrac{1}{C}\sum_{i_j=1}^{C}\left\Vert \nabla f_{i_j}(x_{i_j}^{j-\tau_j})\right\Vert^2\right] + \dfrac{\eta L^2}{CN_m}\E\left[\delta^{t-\tau_t}\right]
\end{align*}

\begin{align*}
    \E\left[ f(x^{t+1}) \right] \leq &\E\left[ f(x^{t}) \right] - \dfrac{\eta}{2CN_m}\E\left[ \left\Vert \nabla f(x^t) \right\Vert^2 \right]  + \left(\dfrac{\eta^2L}{C^2N_m} - \dfrac{\eta}{2CN_m}\right)\E\left[\left\Vert\dfrac{1}{C}\sum_{i_t=1}^{C} \nabla f_{i_t}(x_{i_t}^{t-\tau_t})\right\Vert^2\right]\\ &+ \dfrac{\eta^3L^2\tau_t}{C^3N_{m}^{2}}\sum_{j=t-\tau_t}^{t-1}\E\left[\dfrac{1}{C}\sum_{i_j=1}^{C}\left\Vert \nabla f_{i_j}(x_{i_j}^{j-\tau_j})\right\Vert^2\right] + \left(\dfrac{6\eta^2L^3}{C^2N_m}+\dfrac{\eta L^2}{CN_m}\right)\E\left[\delta^{t-\tau_t}\right]\\ &+ \left(\dfrac{\eta^2L}{2C^2N_m} + \dfrac{\eta^3L^2\tau_{t}^{2}}{C^3N_{m}^{2}}\right)\sigma^2 + \dfrac{4\eta^2Ls^2}{C^2N_m}
\end{align*}

Applying summation from $t=0$ to $K-1$:

\begin{align*}
    \E\left[ f(x^{K}) \right] \leq &\E\left[ f(x^{0}) \right] - \dfrac{\eta}{2CN_m}\sum_{t=0}^{K-1}\E\left[ \left\Vert \nabla f(x^t) \right\Vert^2 \right]\\ &+ \left(\dfrac{\eta^2L}{C^2N_m} - \dfrac{\eta}{2CN_m}\right)\sum_{t=0}^{K-1}\E\left[\left\Vert\dfrac{1}{C}\sum_{i_t=1}^{C} \nabla f_{i_t}(x_{i_t}^{t-\tau_t})\right\Vert^2\right]\\ &+ \underbrace{\dfrac{\eta^3L^2\tau_t}{C^3N_{m}^{2}}\sum_{t=0}^{K-1}\tau_t\sum_{j=t-\tau_t}^{t-1}\E\left[\dfrac{1}{C}\sum_{i_j=1}^{C}\left\Vert \nabla f_{i_j}(x_{i_j}^{j-\tau_j})\right\Vert^2\right]}_{T_5} + \left(\dfrac{6\eta^2L^3}{C^2N_m}+\dfrac{\eta L^2}{CN_m}\right)\sum_{t=0}^{K-1}\E\left[\delta^{t-\tau_t}\right]\\ &+ \dfrac{K\eta^2L\sigma^2}{2C^2N_m} + \dfrac{\eta^3L^2\sigma^2}{C^3N_{m}^{2}}\sum_{t=0}^{K-1}\tau_{t}^{2} + \dfrac{4\eta^2Ls^2K}{C^2N_m}
\numberthis \label{eqn23}
\end{align*}

Expanding the term $T_5$ using upper bound of $\tau_t$ i.e. $T$

\begin{align*}
    T_5 =& \dfrac{\eta^3L^2\tau_t}{C^3N_{m}^{2}}\sum_{t=0}^{K-1}\tau_t\sum_{j=t-\tau_t}^{t-1}\E\left[\dfrac{1}{C}\sum_{i_j=1}^{C}\left\Vert \nabla f_{i_j}(x_{i_j}^{j-\tau_j})\right\Vert^2\right]\\ =& \dfrac{\eta^3L^2T^2}{C^3N_{m}^{2}}\sum_{t=0}^{K-1}\E\left[\dfrac{1}{C}\sum_{i_t=1}^{C}\left\Vert \nabla f_{i_t}(x_{i_t}^{t-\tau_t})\right\Vert^2\right]\\ =& \dfrac{\eta^3L^2T^2}{C^3N_{m}^{2}}\sum_{t=0}^{K-1}\left(12L^2\E\left[\delta^{t-\tau_t}\right] + 8s^2 + 2\E\left[\left\Vert\dfrac{1}{C}\sum_{i_t=1}^{C} \nabla f_{i_t}(x_{i_t}^{t-\tau_t}) \right\Vert^2\right]\right)\\ =& \dfrac{12\eta^3L^4T^2}{C^3N_{m}^{2}}\sum_{t=0}^{K-1}\E\left[\delta^{t-\tau_t}\right] + \dfrac{8\eta^3L^2T^2s^2K}{C^3N_{m}^{2}} + \dfrac{2\eta^3L^2T^2}{C^3N_{m}^{2}}\sum_{t=0}^{K-1}\E\left[\left\Vert\dfrac{1}{C}\sum_{i_t=1}^{C} \nabla f_{i_t}(x_{i_t}^{t-\tau_t}) \right\Vert^2\right]
\numberthis \label{eqn24}
\end{align*}

Plugging back equation \ref{eqn24} into equation \ref{eqn23} we get:

\begin{align*}
    \E\left[ f(x^{K}) \right] \leq &\E\left[ f(x^{0}) \right] - \dfrac{\eta}{2CN_m}\sum_{t=0}^{K-1}\E\left[ \left\Vert \nabla f(x^t) \right\Vert^2 \right]\\  &+ \left(\dfrac{\eta^2L}{C^2N_m} - \dfrac{\eta}{2CN_m} + \dfrac{2\eta^3L^2T^2}{C^3N_{m}^{2}}\right)\sum_{t=0}^{K-1}\E\left[\left\Vert\dfrac{1}{C}\sum_{i_t=1}^{C} \nabla f_{i_t}(x_{i_t}^{t-\tau_t})\right\Vert^2\right]\\ &+ \left(\dfrac{6\eta^2L^3}{C^2N_m}+\dfrac{\eta L^2}{CN_m} + \dfrac{12\eta^3L^4T^2}{C^3N_{m}^{2}}\right)\sum_{t=0}^{K-1}\E\left[\delta^{t-\tau_t}\right] + \dfrac{K\eta^2L\sigma^2}{2C^2N_m}\\ &+ \dfrac{\eta^3L^2T^2\sigma^2K}{C^3N_{m}^{2}} + \dfrac{4\eta^2Ls^2K}{C^2N_m} + \dfrac{8\eta^3L^2T^2s^2K}{C^3N_{m}^{2}}
\end{align*}

Setting $A_1 > 0$ and using the result from lemma \ref{lemma2}

\begin{align*}
    \E\left[ f(x^{K}) \right] \leq &\E\left[ f(x^{0}) \right] - \dfrac{\eta}{2CN_m}\sum_{t=0}^{K-1}\E\left[ \left\Vert \nabla f(x^t) \right\Vert^2 \right]\\  &+ \left(\dfrac{\eta^2L}{C^2N_m} - \dfrac{\eta}{2CN_m} + \dfrac{2\eta^3L^2T^2}{C^3N_{m}^{2}}\right)\sum_{t=0}^{K-1}\E\left[\left\Vert\dfrac{1}{C}\sum_{i_t=1}^{C} \nabla f_{i_t}(x_{i_t}^{t-\tau_t})\right\Vert^2\right]\\ &+ \left(\dfrac{6\eta^2L^3}{C^2N_m}+\dfrac{\eta L^2}{CN_m} + \dfrac{12\eta^3L^4T^2}{C^3N_{m}^{2}}\right)\left[
\begin{aligned}
&\dfrac{2\eta^2K}{A_1C}\left[\Delta_{a}^{2} + \dfrac{8es^2(\kappa-1)^2}{CN_m} + \dfrac{e\sigma^2(\kappa-1)}{CN_m}\right]\\ &+ \dfrac{4\eta^2e(\kappa-1)^2}{A_1C^2N_m}\sum_{t=0}^{K-1}\E\left[\left\Vert \dfrac{1}{C}\sum_{i_t=1}^{C} \nabla f_{i_t}(x_{i_t}^{t-\tau_t}) \right\Vert^2\right]
\end{aligned}
\right]\\ &+ \dfrac{K\eta^2L\sigma^2}{2C^2N_m} + \dfrac{\eta^3L^2T^2\sigma^2K}{C^3N_{m}^{2}} + \dfrac{4\eta^2Ls^2K}{C^2N_m} + \dfrac{8\eta^3L^2T^2s^2K}{C^3N_{m}^{2}}
\end{align*}

Setting $\dfrac{2\eta LT^2}{CN_m}\leq 1$ , we get:

\begin{align*}
    \dfrac{\eta}{2CN_m}\sum_{t=0}^{K-1}\E\left[ \left\Vert \nabla f(x^t) \right\Vert^2 \right] \leq &\E\left[ f(x^{0}) \right] - \E\left[ f(x^{K}) \right]\\ &+ A_2 \sum_{t=0}^{K-1}\E\left[\left\Vert\dfrac{1}{C}\sum_{i_t=1}^{C} \nabla f_{i_t}(x_{i_t}^{t-\tau_t})\right\Vert^2\right]\\ &+ \left(\dfrac{12\eta^2L^3}{C^2N_m} + \dfrac{\eta L^2}{CN_m}\right)\left(\dfrac{2\eta^2K}{A_1C}\left[\Delta_{a}^{2} + \dfrac{8es^2(\kappa-1)^2}{CN_m} + \dfrac{e\sigma^2(\kappa-1)}{CN_m}\right]\right)\\ &+ \dfrac{K\eta^2L\sigma^2}{C^2N_m} + \dfrac{8\eta^2Ls^2K}{C^2N_m}  
\end{align*}

Finally, on setting $A_2 \leq 0$ and $A_3 \leq 1$ we get the main inequality:

\begin{align*}
    \dfrac{1}{K}\sum_{t=0}^{K-1}\E\left[ \left\Vert \nabla f(x^t) \right\Vert^2 \right] \leq &\dfrac{2CN_m}{\eta K}\left(f(x^{0}) - f(x^*)\right) + \dfrac{2\eta}{C}\left(2N_m + L\left(\sigma^2 + 8s^2\right)\right)
\end{align*}

\section{Proof for Corollary \ref{corollary1.1}}
\label{appendixcorollaryproof}

We start by restating all the constraints:

\begin{enumerate}
    \item $\dfrac{2\eta LT^2}{CN_m} \leq 1 \implies \eta \leq \dfrac{CN_m}{2LT^2}$
    \item $A_2\leq 0$
        \begin{align*}
        \dfrac{\eta^2L}{C^2N_m} - \dfrac{\eta}{2CN_m} + \dfrac{2\eta^3L^2T^2}{C^3N_{m}^{2}} + \dfrac{4\eta^2e(\kappa-1)^2}{A_1C^2N_m}\left[\dfrac{6\eta^2L^3}{C^2N_m}+\dfrac{\eta L^2}{CN_m} + \dfrac{12\eta^3L^4T^2}{C^3N_{m}^{2}}\right] \leq 0
        \end{align*}
        Taking $\dfrac{\eta}{CN_m}$ common and $A_1\geq \dfrac{1}{2}$
        \begin{align*}
        \implies\dfrac{\eta L}{C} + \dfrac{2\eta^2L^2T^2}{C^2N_m} + \dfrac{48\eta^3L^3e(\kappa-1)^2}{C^3N_m} + \dfrac{8\eta^2 L^2e(\kappa-1)^2}{C^2N_m} + \dfrac{96\eta^4L^4T^2e(\kappa-1)^2}{C^4N_{m}^{2}}\leq \dfrac{1}{2}
        \end{align*}
        We set each of the terms on the LHS smaller than $\dfrac{1}{10}$ to get the conditions on $\eta$.
        \begin{itemize}
            \item $\dfrac{\eta L}{C}\leq \dfrac{1}{10} \implies \eta \leq \dfrac{C}{10L}$
            \item $\dfrac{2\eta^2L^2T^2}{C^2N_m} \leq \dfrac{1}{10} \implies \eta \leq \dfrac{CN_{m}^{\frac{1}{2}}}{LT\sqrt{20}}$
            \item $\dfrac{48\eta^3L^3e(\kappa-1)^2}{C^3N_m} \leq \dfrac{1}{10} \implies \eta \leq \dfrac{CN_{m}^{\frac{1}{3}}}{Le^{\frac{1}{3}}(\kappa-1)^{\frac{2}{3}}(480)^{\frac{1}{3}}}$
            \item
            $\dfrac{8\eta^2 L^2e(\kappa-1)^2}{C^2N_m} \leq \dfrac{1}{10} \implies \eta \leq \dfrac{CN_{m}^{\frac{1}{2}}}{Le^{\frac{1}{2}}(\kappa-1)\sqrt{80}}$
            \item $\dfrac{96\eta^4L^4T^2e(\kappa-1)^2}{C^4N_{m}^{2}}\leq \dfrac{1}{10} \implies \eta \leq \dfrac{CN_{m}^{\frac{1}{2}}}{LT^{\frac{1}{2}}e^{\frac{1}{4}}(\kappa-1)^{\frac{1}{2}}(960)^{\frac{1}{4}}}$
        \end{itemize}
    \item $A_3 \leq 1$
    \begin{align*}
    \dfrac{1}{A_1}\left(\dfrac{12\eta^2L^3}{C^2N_m} + \dfrac{\eta L^2}{CN_m}\right)\left(\Delta_{a}^{2} + \dfrac{8es^2(\kappa-1)^2}{CN_m} + \dfrac{e\sigma^2(\kappa-1)}{CN_m}\right)\leq 1
    \end{align*}
    \begin{align*}
    \implies 1 \geq& \dfrac{24\eta^2L^3\Delta_{a}^{2}}{CN_m} + \dfrac{24\eta^2L^3e(\kappa-1)(8s^2(\kappa-1)+\sigma^2)}{C^2N_{m}^{2}} + \dfrac{2\eta L^2\Delta_{a}^{2}}{N_m}\\ &+ \dfrac{2\eta L^2e(\kappa-1)(8s^2(\kappa-1)+\sigma^2)}{CN_{m}^{2}}
    \end{align*}
    We set each of the terms on the LHS smaller than $\dfrac{1}{4}$ to get the conditions on $\eta$.
    \begin{itemize}
        \item $\dfrac{24\eta^2L^3\Delta_{a}^{2}}{CN_m} \leq \dfrac{1}{4} \implies \eta \leq \dfrac{C^{\frac{1}{2}}N_{m}^{\frac{1}{2}}}{4\sqrt{6}L^{\frac{3}{2}}\Delta_{a}}$
        \item 
        $\dfrac{24\eta^2L^3e(\kappa-1)(8s^2(\kappa-1)+\sigma^2)}{C^2N_{m}^{2}}\leq \dfrac{1}{4} \implies \eta \leq \dfrac{CN_m}{4\sqrt{6}L^{\frac{3}{2}}e^{\frac{1}{2}}(\kappa-1)^{\frac{1}{2}}(8s^2(\kappa-1)+\sigma^2)^{\frac{1}{2}}}$
        \item
        $\dfrac{2\eta L^2\Delta_{a}^{2}}{N_m}\leq \dfrac{1}{4} \implies \eta \leq \dfrac{N_m}{8L^2\Delta_{a}^{2}}$
        \item
        $\dfrac{2\eta L^2e(\kappa-1)(8s^2(\kappa-1)+\sigma^2)}{CN_{m}^{2}} \leq \dfrac{1}{4} \implies \eta \leq \dfrac{CN_{m}^{2}}{8L^2e(\kappa-1)(8s^2(\kappa-1)+\sigma^2)}$
    \end{itemize}
\end{enumerate}

Next we assume $\eta = C\sqrt{\dfrac{N_m}{K[2N_m+L(\sigma^2 + 8s^2)]}}$ and reduce the constraints obtained above:
\begin{align*}
C\sqrt{\dfrac{N_m}{K[2N_m+L(\sigma^2 + 8s^2)]}} \leq \dfrac{1}{2L} \min
\left\{
\begin{aligned}
& \dfrac{CN_{m}^{\frac{1}{2}}}{\sqrt{5}T^2},\dfrac{C}{5}, \dfrac{CN_{m}^{\frac{1}{3}}}{2\sqrt{5}e^{\frac{1}{2}}(\kappa-1)}, \dfrac{CN_{m}^{\frac{1}{2}}}{3T^{\frac{1}{2}}e^{\frac{1}{4}}(\kappa-1)^{\frac{1}{2}}},\\ &\dfrac{C^{\frac{1}{2}}N_{m}^{\frac{1}{2}}}{2\sqrt{6}L^{\frac{1}{2}}\Delta_{a}}, \dfrac{CN_m}{2\sqrt{6}L^{\frac{1}{2}}e^{\frac{1}{2}}(\kappa-1)^{\frac{1}{2}}(8s^2(\kappa-1)+\sigma^2)^{\frac{1}{2}}}, \\ &\dfrac{N_m}{4L\Delta_{a}^{2}}, \dfrac{CN_{m}^{2}}{4Le(\kappa-1)(8s^2(\kappa-1)+\sigma^2)}
\end{aligned}
\right\}
\end{align*}

Upon simplifying the above expression for $K$, we get:
\begin{align*}
K \geq \dfrac{4L^2}{2N_m + L(\sigma^2 + 8s^2)} \max
\left\{
\begin{aligned}
&5T^4, 25N_m, 20e(\kappa -1)^2 N_m^{\frac{1}{3}}, \\ &9T\sqrt{e}(\kappa-1),  24L\Delta_a^2C, \\ &\dfrac{24Le(\kappa-1)(\sigma^2+8s^2(\kappa-1))}{N_m},  \dfrac{16L^2\Delta_a^2C^2}{N_m}, \\ &\dfrac{16L^2e^2(\kappa-1)^2(\sigma^2+8s^2(\kappa-1))^2}{N_m^3} 
\end{aligned}
\right\}
\end{align*}

Now we substitute $\eta$ back into the main inequality (\ref{eqn12}) and simplify to get:
\begin{align*}
\dfrac{1}{K} \sum_{t=0}^{K-1} \E\left[ \left\Vert \nabla f(x^t) \right\Vert^2 \right] \leq \dfrac{2(f(x^0) - f(x^*))}{\sqrt{K}} \sqrt{ 2N_m^2 + LN_m(\sigma^2 + 8s^2)}
\end{align*}

\section{Proofs for Supporting Lemmas}
\label{appendixsupportinglemmas}

\begin{lemma}
\label{lemma1}
\begin{align*}
    \E\left[\dfrac{1}{C}\sum_{i_t=1}^{C}\left\Vert \nabla f_{i_t}(x_{i_t}^{t-\tau_t}) \right\Vert^2\right] \leq & 12L^2\E\left[\delta^{t-\tau_t}\right] + 8s^2\\  &+ 2\E\left[\left\Vert\dfrac{1}{C}\sum_{j_t=1}^{C} \nabla f_{j_t}(x_{j_t}^{t-\tau_t}) \right\Vert^2\right]
\end{align*}

Proof. The LHS can be bounded by:

\begin{align*}
    \E\left[\dfrac{1}{C}\sum_{i_t=1}^{C}\left\Vert \nabla f_{i_t}(x_{i_t}^{t-\tau_t}) \right\Vert^2\right] \leq &\underbrace{2\E\left[\dfrac{1}{C}\sum_{i_t=1}^{C}\left\Vert \nabla f_{i_t}(x_{i_t}^{t-\tau_t}) - \dfrac{1}{C}\sum_{j_t=1}^{C}\nabla f_{j_t}(x_{j_t}^{t-\tau_t}) \right\Vert^2\right]}_{D_1}\\ &+ 2\E\left[\left\Vert\dfrac{1}{C}\sum_{j_t=1}^{C}\nabla f_{j_t}(x_{j_t}^{t-\tau_t}) \right\Vert^2\right]
\numberthis \label{eqn25}
\end{align*}

Expanding the term $D_1$:

\begin{align*}
    D_1 =& 2\E\left[\dfrac{1}{C}\sum_{i_t=1}^{C}\left\Vert \nabla f_{i_t}(x_{i_t}^{t-\tau_t}) - \dfrac{1}{C}\sum_{j_t=1}^{C}\nabla f_{j_t}(x_{j_t}^{t-\tau_t}) \right\Vert^2\right]\\ \leq & 2\E\left[\dfrac{1}{C}\sum_{i_t=1}^{C}\left\Vert \nabla f_{i_t}(x_{i_t}^{t-\tau_t}) - \nabla f(x^{t-\tau_t}) - \dfrac{1}{C}\sum_{j_t=1}^{C}\nabla f_{j_t}(x_{j_t}^{t-\tau_t}) + \nabla f(x^{t-\tau_t}) \right\Vert^2\right]\\ \leq &4\E\left[\dfrac{1}{C}\sum_{i_t=1}^{C}\left\Vert \nabla f_{i_t}(x_{i_t}^{t-\tau_t}) - \nabla f(x^{t-\tau_t})\right\Vert^2\right] + 4\E\left[\dfrac{1}{C}\sum_{i_t=1}^{C}\left\Vert \nabla f(x^{t-\tau_t}) - \dfrac{1}{C}\sum_{j_t=1}^{C}\nabla f_{j_t}(x_{j_t}^{t-\tau_t}) \right\Vert^2\right]\\ \leq & 4\E\left[\dfrac{1}{C}\sum_{i_t=1}^{C}\left\Vert \nabla f_{i_t}(x_{i_t}^{t-\tau_t}) - \nabla f_{i_t}(x^{t-\tau_t}) + \nabla f_{i_t}(x^{t-\tau_t}) - \nabla f(x^{t-\tau_t})\right\Vert^2\right]\\ & + 4\E\left[\dfrac{1}{C}\sum_{i_t=1}^{C}\left\Vert \dfrac{1}{C} \sum_{j_t=1}^{C}\left(\nabla f_{j_t}(x^{t-\tau_t}) - \nabla f_{j_t}(x_{j_t}^{t-\tau_t}) \right)\right\Vert^2\right]\\ \leq & 8\E\left[\dfrac{1}{C}\sum_{i_t=1}^{C}\left\Vert \nabla f_{i_t}(x_{i_t}^{t-\tau_t}) - \nabla f_{i_t}(x^{t-\tau_t})\right\Vert^2\right] + 8\E\left[\dfrac{1}{C}\sum_{i_t=1}^{C}\left\Vert \nabla f_{i_t}(x^{t-\tau_t}) - \nabla f(x^{t-\tau_t})\right\Vert^2\right]\\ &+ 4\E\left[\dfrac{1}{C}\sum_{i_t=1}^{C}\left\Vert \dfrac{1}{C} \sum_{j_t=1}^{C}\left(\nabla f_{j_t}(x^{t-\tau_t}) - \nabla f_{j_t}(x_{j_t}^{t-\tau_t}) \right)\right\Vert^2\right]\\ \leq & 8\E\left[\dfrac{1}{C}\sum_{i_t=1}^{C}\left\Vert \nabla f_{i_t}(x_{i_t}^{t-\tau_t}) - \nabla f_{i_t}(x^{t-\tau_t})\right\Vert^2\right] + 8\E\left[\dfrac{1}{C}Cs^2\right]\\ &+ 4\E\left[\dfrac{1}{C}C\left\Vert \dfrac{1}{C} \sum_{j_t=1}^{C}\left(\nabla f_{j_t}(x^{t-\tau_t}) - \nabla f_{j_t}(x_{j_t}^{t-\tau_t}) \right)\right\Vert^2\right]
    \end{align*}
    \begin{align*}
    D_1 \leq & 8\E\left[\dfrac{1}{C}\sum_{i_t=1}^{C}\left\Vert \nabla f_{i_t}(x_{i_t}^{t-\tau_t}) - \nabla f_{i_t}(x^{t-\tau_t})\right\Vert^2\right] + 8s^2\\ &+ 4\E\left[\dfrac{1}{C} \sum_{j_t=1}^{C}\left\Vert\nabla f_{j_t}(x^{t-\tau_t}) - \nabla f_{j_t}(x_{j_t}^{t-\tau_t}) \right\Vert^2\right]\\ \leq & 12\E\left[\dfrac{1}{C}\sum_{i_t=1}^{C}\left\Vert \nabla f_{i_t}(x_{i_t}^{t-\tau_t}) - \nabla f_{i_t}(x^{t-\tau_t})\right\Vert^2\right] + 8s^2\\ \leq & 12L^2\E\left[\underbrace{\dfrac{1}{C}\sum_{i_t=1}^{C}\left\Vert x_{i_t}^{t-\tau_t} - x^{t-\tau_t}\right\Vert^2}_{\delta^{t-\tau_t}}\right] + 8s^2\\ \leq & 12L^2\E\left[\delta^{t-\tau_t}\right] + 8s^2 
\numberthis \label{eqn26}
\end{align*}

Substituting back equation (\ref{eqn26}) into equation (\ref{eqn25}), we get:

\begin{align*}
    \E\left[\dfrac{1}{C}\sum_{i_t=1}^{C}\left\Vert \nabla f_{i_t}(x_{i_t}^{t-\tau_t}) \right\Vert^2\right] \leq & 12L^2\E\left[\delta^{t-\tau_t}\right] + 8s^2 + 2\E\left[\left\Vert\dfrac{1}{C}\sum_{j_t=1}^{C} \nabla f_{j_t}(x_{j_t}^{t-\tau_t}) \right\Vert^2\right]        
\end{align*}

\end{lemma}

\begin{lemma}
\label{lemma2}

\begin{align*}
    \sum_{t=0}^{K-1}\E\left[\delta^{t-\tau_t}\right] \leq & \dfrac{2\eta^2K}{A_1C}\left[\Delta_{a}^{2} + \dfrac{8es^2(\kappa-1)^2}{CN_m} + \dfrac{e\sigma^2(\kappa-1)}{CN_m}\right]\\ &+ \dfrac{4\eta^2e(\kappa-1)^2}{A_1C^2N_m}\sum_{t=0}^{K-1}\E\left[\left\Vert \dfrac{1}{C}\sum_{i_t=1}^{C} \nabla f_{i_t}(x_{i_t}^{t-\tau_t}) \right\Vert^2\right] \quad \text{where } A_1 > 0
\end{align*}

Proof. 

\begin{align*}
    \E\left[\delta^{t-\tau_t}\right] =& \E\left[\dfrac{1}{C}\sum_{i_t=1}^{C}\left\Vert x_{i_t}^{t-\tau_t} - x^{t-\tau_t} \right\Vert^2\right]
\end{align*}

Let $p = t-\tau_t$ such that $p=a\kappa + t^\prime$ for $t^\prime \in [0,\kappa)$ where $\kappa$ represents the communication period across cluster-scale.

\begin{align*}
    \E\left[\delta^p\right] =& \E\left[\dfrac{1}{C}\sum_{i_t=1}^{C}\left\Vert x_{i_t}^{p} - x^{p} \right\Vert^2\right]\\ \leq& \dfrac{1}{C}\sum_{i_t=1}^{C} \E\left[\left\Vert x_{i_t}^{p} - x^{a\kappa}\right\Vert^2\right]\\ =& \dfrac{1}{C}\sum_{i_t=1}^{C}\E\left[\left\Vert x_{i_t}^{a\kappa} - x^{a\kappa} - \dfrac{\eta}{C}\sum_{j=a\kappa}^{p-1}S_{n_{i_j,j}}g_{i_j}^{j-\tau_j}\right\Vert^2\right]
    \end{align*}
    \begin{align*}
    \E\left[\delta^p\right] \leq& \dfrac{2}{C}\sum_{i_t=1}^{C} \E\left[\left\Vert x_{i_t}^{a\kappa} - x^{a\kappa}\right\Vert^2\right] + \dfrac{2\eta^2}{C}\sum_{i_t=1}^{C}\E\left[\left\Vert \dfrac{1}{C}\sum_{j=a\kappa}^{p-1}S_{n_{i_j,j}}g_{i_j}^{j-\tau_j} \right\Vert^2\right]\\ \leq& \dfrac{2}{C}\sum_{i_t=1}^{C} \E\left[\left\Vert x_{i_t}^{a\kappa} - x^{a\kappa}\right\Vert^2\right] + \underbrace{\dfrac{2\eta^2}{C^2N_m}\sum_{i_t=1}^{C}\E\left[\sum_{i_j=1}^{C}\left\Vert \dfrac{1}{C}\sum_{j=a\kappa}^{p-1}g_{i_j}^{j-\tau_j} \right\Vert^2\right]}_{D_2}
\numberthis \label{eqn27}
\end{align*}

Expanding the term $D_2$:

\begin{align*}
    D_2 =& \dfrac{2\eta^2}{C^2N_m}\sum_{i_t=1}^{C}\E\left[\sum_{i_j=1}^{C}\left\Vert \dfrac{1}{C}\sum_{j=a\kappa}^{p-1}g_{i_j}^{j-\tau_j} \right\Vert^2\right]\\ =& \dfrac{2\eta^2}{C^3N_m}\sum_{i_j=1}^{C}\E\left[\left\Vert \sum_{j=a\kappa}^{p-1}g_{i_j}^{j-\tau_j} \right\Vert^2\right]\\ \leq& \dfrac{2\eta^2}{C^3N_m}e(p-a\kappa)\sum_{i_j=1}^{C}\sum_{j=a\kappa}^{p-1}\E\left[\left\Vert \nabla f_{i_j}(x_{i_j}^{j-\tau_j}) \right\Vert^2\right]\\ &+ \dfrac{2\eta^2}{C^3N_m}e\sum_{i_j=1}^{C}\sum_{j=a\kappa}^{p-1}\E\left[\left\Vert g_{i_j}^{j-\tau_j} - \nabla f_{i_j}(x_{i_j}^{j-\tau_j})  \right\Vert^2\right]\\ \leq& \dfrac{2\eta^2e(p-a\kappa)}{C^2N_m}\sum_{j=a\kappa}^{p-1}\E\left[\dfrac{1}{C}\sum_{i_j=1}^{C}\left\Vert \nabla f_{i_j}(x_{i_j}^{j-\tau_j}) \right\Vert^2\right]\\ &+ \dfrac{2\eta^2e(p-a\kappa)}{C^2N_m}\sigma^2 
\end{align*}

Using lemma \ref{lemma1}

\begin{align*}
    D_2 \leq& \dfrac{24\eta^2L^2e(\kappa-1)}{C^2N_m}\sum_{j=a\kappa}^{p-1} \E\left[\delta^{j-\tau_j}\right] + \dfrac{16\eta^2es^2(\kappa-1)^2}{C^2N_m}\\ &+ \dfrac{4\eta^2e(\kappa-1)}{C^2N_m}\sum_{j=a\kappa}^{p-1}\E\left[\left\Vert \dfrac{1}{C}\sum_{i_j=1}^{C}\nabla f_{i_j}(x_{i_j}^{j-\tau_j}) \right\Vert^2\right] + \dfrac{2\eta^2e(\kappa-1)\sigma^2}{C^2N_m}
\numberthis \label{eqn28}
\end{align*}

Substituting back equation (\ref{eqn28}) into equation (\ref{eqn27}):

\begin{align*}
    \E\left[\delta^p\right] \leq& \dfrac{2}{C}\sum_{i_t=1}^{C} \E\left[\left\Vert x_{i_t}^{a\kappa} - x^{a\kappa}\right\Vert^2\right] + \dfrac{24\eta^2L^2e(\kappa-1)}{C^2N_m}\sum_{j=a\kappa}^{p-1} \E\left[\delta^{j-\tau_j}\right] + \dfrac{16\eta^2es^2(\kappa-1)^2}{C^2N_m}\\ &+ \dfrac{4\eta^2e(\kappa-1)}{C^2N_m}\sum_{j=a\kappa}^{p-1}\E\left[\left\Vert \dfrac{1}{C}\sum_{i_j=1}^{C}\nabla f_{i_j}(x_{i_j}^{j-\tau_j}) \right\Vert^2\right] + \dfrac{2\eta^2e(\kappa-1)\sigma^2}{C^2N_m}
\end{align*}

By using assumption \ref{Assumptions}:4, we get:

\begin{align*}
    \E\left[\delta^p\right] \leq& \dfrac{2\eta^2\Delta_{a}^{2}}{C} + \dfrac{24\eta^2L^2e(\kappa-1)}{C^2N_m}\sum_{j=a\kappa}^{p-1} \E\left[\delta^{j-\tau_j}\right] + \dfrac{16\eta^2es^2(\kappa-1)^2}{C^2N_m}\\ &+ \dfrac{4\eta^2e(\kappa-1)}{C^2N_m}\sum_{j=a\kappa}^{p-1}\E\left[\left\Vert \dfrac{1}{C}\sum_{i_j=1}^{C}\nabla f_{i_j}(x_{i_j}^{j-\tau_j}) \right\Vert^2\right] + \dfrac{2\eta^2e(\kappa-1)\sigma^2}{C^2N_m}
\end{align*}

Taking summation from $t=0$ to $K-1$:

\begin{align*}
    \sum_{t=0}^{K-1}\E\left[\delta^{t-\tau_t}\right] \leq& \dfrac{2\eta^2\Delta_{a}^{2}K}{C} + \dfrac{24\eta^2L^2e(\kappa-1)^2}{C^2N_m}\sum_{t=0}^{K-1} \E\left[\delta^{j-\tau_j}\right] + \dfrac{16\eta^2es^2(\kappa-1)^2K}{C^2N_m}\\ &+ \dfrac{4\eta^2e(\kappa-1)^2}{C^2N_m}\sum_{t=0}^{K-1}\E\left[\left\Vert \dfrac{1}{C}\sum_{i_t=1}^{C}\nabla f_{i_t}(x_{i_t}^{t-\tau_t}) \right\Vert^2\right] + \dfrac{2\eta^2e\sigma^2(\kappa-1)K}{C^2N_m}
\end{align*}
\begin{align*}
    \implies \underbrace{\left(1-\dfrac{24\eta^2L^2e(\kappa-1)^2}{C^2N_m}\right)}_{A_1}\sum_{t=0}^{K-1}\E\left[\delta^{t-\tau_t}\right] \leq& \dfrac{2\eta^2\Delta_{a}^{2}K}{C} + \dfrac{16\eta^2es^2(\kappa-1)^2K}{C^2N_m}\\ &+ \dfrac{4\eta^2e(\kappa-1)^2}{C^2N_m}\sum_{t=0}^{K-1}\E\left[\left\Vert \dfrac{1}{C}\sum_{i_t=1}^{C}\nabla f_{i_t}(x_{i_t}^{t-\tau_t}) \right\Vert^2\right]\\ &+ \dfrac{2\eta^2e\sigma^2(\kappa-1)K}{C^2N_m} 
\end{align*}

\begin{align*}
    \implies\sum_{t=0}^{K-1}\E\left[\delta^{t-\tau_t}\right] \leq & \dfrac{2\eta^2K}{A_1C}\left[\Delta_{a}^{2} + \dfrac{8es^2(\kappa-1)^2}{CN_m} + \dfrac{e\sigma^2(\kappa-1)}{CN_m}\right]\\ &+ \dfrac{4\eta^2e(\kappa-1)^2}{A_1C^2N_m}\sum_{t=0}^{K-1}\E\left[\left\Vert \dfrac{1}{C}\sum_{i_t=1}^{C} \nabla f_{i_t}(x_{i_t}^{t-\tau_t}) \right\Vert^2\right]
\end{align*}
This completes the proof for Lemma \ref{lemma2}
\end{lemma}

\end{appendix}

\end{document}